\documentclass{article}
\usepackage[final]{neurips_2022}

\usepackage[utf8]{inputenc} 
\usepackage[T1]{fontenc}    
\usepackage{url}            
\usepackage{booktabs}       
\usepackage{amsfonts}       
\usepackage{nicefrac}       
\usepackage{microtype}      
\usepackage{comment}
\usepackage{threeparttable}
\usepackage{multirow}
\usepackage{booktabs}
\usepackage{tablefootnote}
\usepackage{bbding}
\usepackage{graphicx}
\usepackage{amsmath} 
\usepackage{color}
\usepackage{xspace}
\usepackage{amssymb}
\usepackage{float}
\usepackage{subfigure}
\usepackage{enumitem}
\usepackage{savesym}
\savesymbol{Cross}
\usepackage{marvosym}
\usepackage{lipsum}
\usepackage[dvipsnames]{xcolor}
\usepackage[colorlinks,urlcolor=Fuchsia,driverfallback=dvipdfm]{hyperref}

\newcommand{\vlp}{Egocentric VLP}
\newcommand{\dataset}{EgoClip}
\newcommand{\model}{EgoNCE\xspace}
\newcommand{\eval}{EgoMCQ}
\newcommand{\epic}{EPIC-KITCHENS-100}

\newcommand{\mir}{Multi-Instance Retrieval}
\newcommand{\nlq}{Natural Language Query}
\newcommand{\mq}{Moment Query}
\newcommand{\ossc}{Object State Change Classification}
\newcommand{\web}{WebVid-2M}
\newcommand{\ccweb}{CC3M+WebVid-2M}
\newcommand{\howto}{HowTo100M}
\usepackage{colortbl}
\usepackage{wrapfig,lipsum,booktabs}
\definecolor{citecolor}{HTML}{0071bc}

\definecolor{LightCyan}{rgb}{0.73,0.73,0.83}
\definecolor{citecolor1}{HTML}{5DADE2}
\definecolor{citecolor2}{HTML}{A569BD }

\title{Egocentric Video-Language Pretraining}
\author{%
	Kevin Qinghong Lin$^1$,~
	Alex Jinpeng Wang$^1$,~
	Mattia Soldan$^3$,~
	Michael Wray$^2$,\\
\textbf{
	Rui Yan$^1$,~
	Eric Zhongcong Xu$^1$,~
	Difei Gao$^1$,~
	Rongcheng Tu$^4$,}\\
\textbf{
	Wenzhe Zhao$^4$,~
	Weijie Kong$^4$,~
	Chengfei Cai$^4$,~
	Hongfa Wang$^4$,}\\
\textbf{
    Dima Damen$^2$,~
	Bernard Ghanem$^3$,~
	Wei Liu$^4$,~
	and~Mike Zheng Shou$^1$\textsuperscript{\Letter}}
	\\\\
	$^1$Show Lab, National University of Singapore\quad
	$^2$University of Bristol\\
	$^3$King Abdullah University of Science and Technology\quad 
	$^4$Tencent Data Platform
}
            
\begin{document}

\maketitle
\newcommand\blfootnote[1]{%
  \begingroup
  \renewcommand\thefootnote{}\footnote{#1}%
  \addtocounter{footnote}{-1}%
  \endgroup
}

\begin{abstract}
Video-Language Pretraining~(VLP), which aims to learn transferable representation to advance a wide range of video-text downstream tasks, has recently received increasing attention. 
Best performing works rely on large-scale, 3rd-person video-text datasets, such as HowTo100M.
In this work, we exploit the recently released Ego4D dataset to pioneer \vlp~along three directions.
(i)~We create \dataset, a 1st-person video-text pretraining dataset comprising 3.8M clip-text pairs well-chosen from Ego4D, covering a large variety of human daily activities.
(ii)~We propose a novel pretraining objective, dubbed \model, which adapts video-text contrastive learning to the egocentric domain by mining egocentric-aware positive and negative samples.
(iii)~We introduce \eval, a development benchmark that is close to \dataset~and hence can support effective validation and fast exploration of our design decisions in \dataset~and \model.
Furthermore, we demonstrate strong performance on five egocentric downstream tasks across three datasets: video-text retrieval on \epic; action recognition on Charades-Ego; natural language query, moment query, and object state change classification on Ego4D challenge benchmarks.
The dataset and code are available at \textcolor{Fuchsia}{\url{https://github.com/showlab/EgoVLP}}.\blfootnote{\Letter: Corresponding Author.}
\end{abstract}

\section{Introduction}
With the recent interest boom in computer vision and natural language processing, Video-Language Pretraining~(VLP) has prevailed, which aims to learn strong and transferable video-language representation for powering a broad spectrum of video-text downstream tasks, such as video-text retrieval~\cite{xu2016msr, patrick2020support, bain2021frozen}, video question answering~\cite{msrvttqamsvdqa,yu2018joint,zhu2020actbert}, and video captioning~\cite{krishna2017dense, wang2018reconstruction, zhou2018end}.
The success of VLP mainly stems from the availability of large-scale open-world video-text datasets~\cite{miech2019howto100m}, which subsume a large number of videos sourced from the Web~(e.g., YouTube) and pair videos with associated textual information.
For instance, \howto~\cite{miech2019howto100m} collects $134$K hours of instructional videos accompanied by noisy narrations yielded from Automatic Speech Recognition~(ASR).
\web~\cite{bain2021frozen} scrapes $2.5$M descriptive videos with well-formed long captions.

Despite reaching an impressive data scale, videos in those existing video-text pretraining datasets are often of 3rd-person views and may have been edited before posting on the Web.
Yet, there is a noticeable domain gap between the existing video-text pretraining datasets and 1st-person view videos such as those videos captured by wearable cameras or smart glasses.
Egocentric video has received increasing interests from the academia~(e.g., activity recognition~\cite{caba2015activitynet}, activity anticipation~\cite{abu2018will}, and video summarization~\cite{ma2002user}) and industry (various applications in robotics and augmented reality).
However, due to such a domain gap, directly transferring the existing VLP models to egocentric downstream tasks cannot fully unleash the potential of large-scale pretraining approaches, which we have confirmed in the later experimental section.
To bridge this gap, we are motivated to develop \vlp~models, which can greatly benefit various egocentric video downstream applications.

However, existing egocentric video datasets are of small scale and domain-specific, making \vlp~prohibitive. As illustrated in Tab.~\ref{dataset}, the formerly largest egocentric video dataset \epic~\cite{kazakos2019epic} focuses on kitchens scenarios and its size is far smaller than those of the 3rd-person pretraining sets \web~\cite{bain2021frozen} and \howto~\cite{miech2019howto100m}.
Fortunately, with the recent introduction of the massive-scale egocentric video dataset Ego4D~\cite{grauman2021ego4d}, it becomes possible to unlock~\vlp.
Ego4D consists of $3,670$ hours of videos with manually annotated narrations from $74$ worldwide locations, covering a large variety of daily-life scenarios and activities.
\begin{table}[t]
\small
\centering
\vspace{-1.5em}
\resizebox{1\textwidth}{!}{
\begin{tabular}{lcccccc}
	\toprule[1pt] 
	\textbf{Dataset} & \textbf{Ego?}    & \textbf{Domain} &  \textbf{Dur~(hrs)}   &   \textbf{\#~Clips} & \textbf{\#~Texts} &  \textbf{Example} \\ 
	\midrule
	MSR-VTT~\cite{xu2016msr}            & \XSolidBrush  &  diverse              & $40$   & $10$K  & $200$K & {\multirow{4}{*}{ \includegraphics[width=0.12\linewidth]{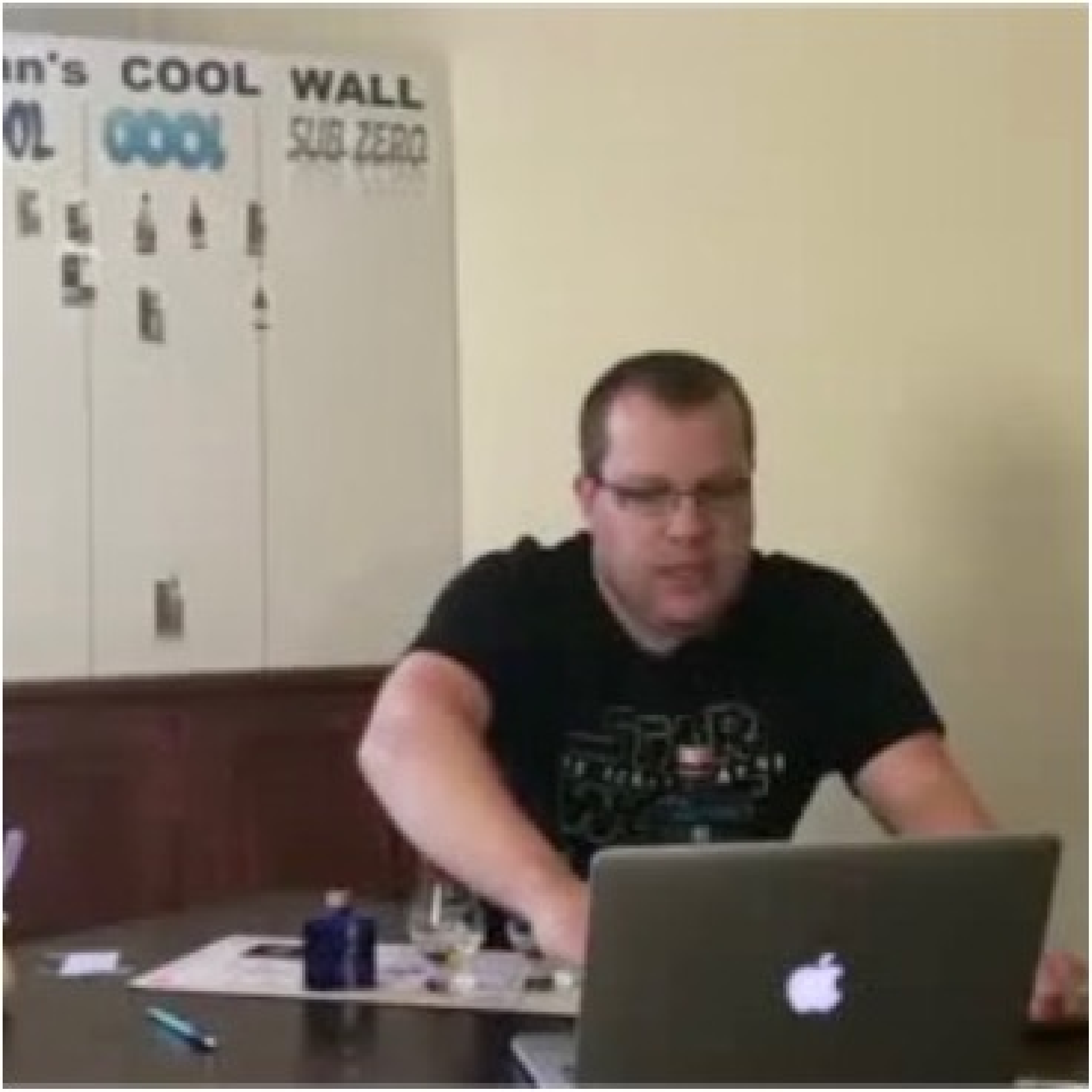} }} \\ 
	YouCook2~\cite{zhou2018towards}     & \XSolidBrush  & cooking               & $176$  & $14$K  & $14$K  &  \\
	ActivityNet Captions~\cite{krishna2017dense} &     \XSolidBrush  &  action  & $849$  & $100$K & $100$K & \\ 
	WebVid-2M~\cite{bain2021frozen}     & \XSolidBrush  & diverse               & $13$K  & $2.5$M & $2.5$M & \\ 
	HowTo100M~\cite{miech2019howto100m} & \XSolidBrush  & instructional         & $134$K & $136$M & $136$M & 3rd-person view \\
	\midrule
	Charades-Ego~\cite{sigurdsson2018charades}  & \Checkmark & home       & $34$  & $30$K & $30$K  &{\multirow{4}{*}{ \includegraphics[width=0.12\linewidth]{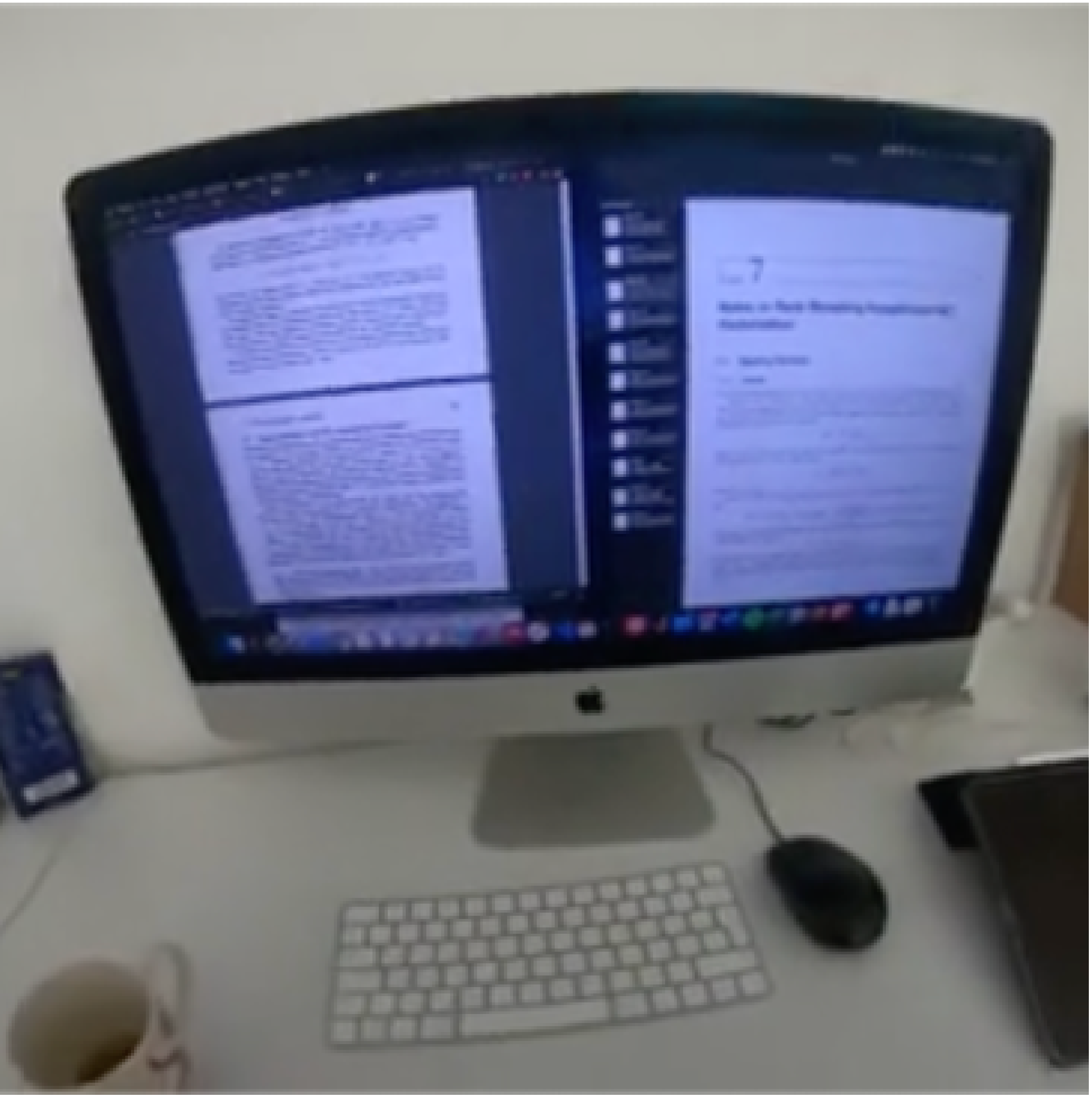} }} \\
	UT-Ego~\cite{lee2012discovering}            & \Checkmark & diverse    & $37$  & $11$K & $11$K  \\
	Disneyworld~\cite{fathi2012social}          & \Checkmark & disneyland & $42$  & $15$K & $15$K  \\
	EPIC-KITCHENS-100~\cite{damen2022rescaling} & \Checkmark & kitchen    & $100$ & $90$K & $90$K  \\
	\textbf{\dataset}                           & \Checkmark & \textbf{diverse} & $\mathbf{2.9K}$ & $\mathbf{3.8M}$ & $\mathbf{3.8M}$ & 1st-person view \\
	\bottomrule[1pt] 
\end{tabular}
}
\centering
\vspace{.5em}
\caption{Comparison of our proposed EgoClip pretraining dataset against the mainstream video-language datasets~(top) and egocentric  datasets~(bottom).}
\label{dataset}
\vspace{-2.5em}
\end{table}

In this work, roused by the favorable scale and diversity of Ego4D, we make a significant effort to pave the way for \vlp~with the following steps:\\
\textbf{(i)} To address the aforementioned issue of lacking a suitable large-scale egocentric video-language pretraining dataset, we create a video-text pretraining dataset \textbf{\dataset} which contains a total of $3.8$M clean 1st-person clip-text pairs selected from Ego4D and covers diverse human daily activities.\\
\textbf{(ii)} To make full use of {\dataset} for video-text representation learning, we propose a novel video-text contrastive objective \textbf{\model} to address unique challenges in egocentric pretraining datasets.\\
\textbf{(iii)} We create a development benchmark i.e., Egocentric Multiple-Choices-Question, dubbed \textbf{\eval}, which contains $39$K questions created from Ego4D and focuses on evaluating video-text alignment. In contrast to other downstream benchmarks, {\eval} has a less discrepancy from \dataset, powering us to accurately validate and quickly iterate our designs of \dataset~and \model.\\
\textbf{(iv)} We conduct extensive experiments to demonstrate the superiority of \vlp~by transferring our pretrained representation to five egocentric downstream benchmarks and achieving state-of-the-art performance: $59.4\%$ nDCG on video-text retrieval of \epic~\cite{kazakos2019epic}~\footnote{\footnotesize \vlp~won championship on Multi-Instance Retrieval, \href{https://epic-kitchens.github.io/2022}{EPIC-Kitchens Challenges @ CVPR 2022}.}, $32.1\%$ mAP on action recognition of Charades-Ego~\cite{sigurdsson2018charades}, and significant boosts over three Ego4D challenges~\footnote{\footnotesize \vlp~won championship on OSCC and 2nd place on NLQ, \href{https://ego4d-data.org/workshops/cvpr22/}{Ego4D Challenges @ CVPR 2022}.}: natural language query, moment query~and object state change classification.
\section{Related Work}
\noindent\textbf{Video-Language Pretraining.}\label{relatedwork_vlp}
The introduction of large-scale video-text datasets~\cite{miech2019howto100m,bain2021frozen} has enabled the emergence of VLP approaches to improve the video-text representation for various vision-language tasks~\cite{anne2017localizing, chen2017sca, msrvttqamsvdqa}, such as MIL-NCE~which \cite{miech2020end} proposes to match clips with multiple captions close in temporal to adapt the video-text misalignment of \howto~\cite{miech2019howto100m}.
Dominant VLP methods can be classified into two groups, namely: joint- and dual-encoders. The former combines videos and texts as a single input to the encoder that performs the multimodal fusion. For instance,~\cite{lei2021less,Sun_2019_ICCV} concatenate videos and texts together before feeding them to a unified transformer.
Conversely, methods like~\cite{bain2021frozen,wang2022object} exploit dual encoders to independently project the video and text inputs into a common space and minimize the distance between the paired representations. 
These approaches are preferred in retrieval settings as they allow for efficient indexing of a single modality~\cite{escorcia2019temporal,miech2021thinking}.
For example, Frozen~\cite{bain2021frozen} employs two separate transformers to encode video and text features and aligns them by video-text InfoNCE~\cite{infonce}.
In our work, we adopt the Frozen~\cite{bain2021frozen} but extend its InfoNCE to \model~via positive and negative sampling for egocentric-friendly pretraining.

\noindent\textbf{Egocentric Video Datasets.}
Egocentric videos, collected by participants using wearable cameras, offer a natural perspective of people's daily activities and raise a range of challenging research topics~\cite{caba2015activitynet, abu2018will, wong2022assistq}.
Several egocentric video datasets have been developed in decades, e.g., \cite{damen2022rescaling, sigurdsson2018charades, li2015delving}.
However, since the collection of egocentric videos is expensive, previous egocentric datasets tend to be small-scale and domain-specific.
These limitations hinder 1st-person view research and fail to match the progress of 3rd-person counterparts, such as VLP~\cite{miech2020end,lei2021less,bain2021frozen}.
Recently, a massive egocentric video dataset Ego4D~\cite{grauman2021ego4d} has been released, which consists of $3,670$ hours of videos collected by $931$ people from $74$ worldwide locations in $9$ different countries, where most videos are accompanied by narrations, audio, 3D meshes, and more.
Furthermore, Ego4D introduces a suite of new challenging benchmarks (e.g., Natural language query and moment query) to fully explore the 1st-person visual experience.
With this step-changing dataset and benchmarks, Ego4D would lead to a new research surge on egocentric visual perception.
\section{\dataset:~An Egocentric Video-Language Pretraining Dataset}

\begin{figure}[!t]
	\centering
	\vspace{-1.5em}
	\includegraphics[width=1.0\linewidth]{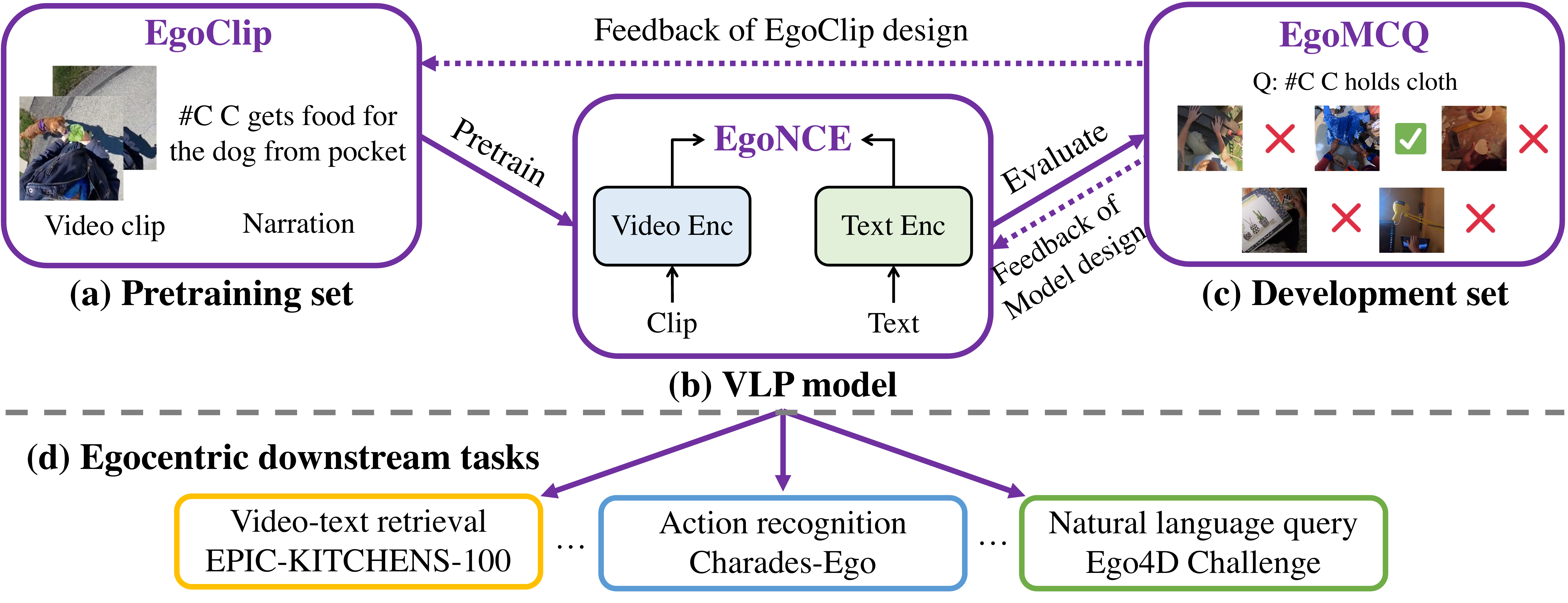}
	\vspace{-1.em}
	\caption{
	Our \vlp~includes: (a) the pretraining set \dataset, (b) the VLP model, and (c) the development set \eval. We use \dataset~to pretrain a VLP model with the \model~ loss and then evaluate on \eval. According to the feedback, we iteratively refine our designs of (a) and (b). We then transfer the pretrained model to downstream tasks relevant to the egocentric domain.}
	\vspace{-1.5em}
	\label{framework}
\end{figure}

\textbf{Data curation.}
For our \dataset{} dataset, we source data from Ego4D~\cite{grauman2021ego4d}, which contains $9,645$ untrimmed videos of varying lengths from $5$ sec to $7$ hrs. From these videos, most are associated with \textit{dense timestamp-level narrations} assigned by two different annotators, describing the camera wearer's activities and interactions with objects. For example, the narration ``{\texttt{\#C C puts the scrapper down.}}'' corresponds to video content that occurred at {$3.70s$}, where  ``\texttt{\#C}'' refers to the camera-wearer.
Notably, narrations in Ego4D are well-aligned with the videos, both temporally and visually.
Prior pretraining datasets are characterized by a much greater level of temporal misalignment between the video and text~(e.g., HowTo100M~\cite{miech2019howto100m} narrations are scraped from ASR, yielding sentences misaligned or even unrelated to video content).
We first filter Ego4D videos with missing narrations~($7.4\%$ of the total video duration) and exclude videos that belong to the validation and test sets of the Ego4D benchmark challenge~\cite{grauman2021ego4d}~(a further $23.9\%$ of the total video duration). Next, we retain textual annotation from both narrators in \dataset, allowing us to consider narration diversity when pairing video and text for pretraining purposes. Finally, we adopt several criteria to filter the video and textual narrations, further reducing noise~(detailed steps are provided in Supplementary~\ref{a1}). Overall, this procedure yields $2.9$K hours of videos with $3.85$ million narrations which cover $2927$ hours of video from $129$ different scenarios. \dataset~has $21.9$ clips per minute with an average clip length of $1.0$ seconds and a standard deviation of $0.9$ seconds (the longest clip is up to $60$s). Additional analyses are included in the Supplementary~\ref{a3}.

\textbf{Creation of clip-text pairs.}
Clip-text pairs are the common data format for VLP, but are usually not present in untrimmed video datasets with only a weak matching between narrations captions and videos.
This was first discussed in \howto~\cite{miech2019howto100m}, which pairs subtitles to video clips with corresponding time intervals to produce noisy pairs.
This is not suitable for Ego4D since each narration is annotated with a single timestamp rather than an interval.
Thus, we design \textit{a contextual variable-length clip pairing strategy}.
Formally, narrations per video in Ego4D are organized as a sequence of sentences $\{\mathcal{T}_0, \cdots, \mathcal{T}_n\}$ with exact timestamps $\{t_0, \cdots, t_n\}$, indicating an event $i$ described by $\mathcal{T}_i$ happened in the moment $t_i$.
For a narration $\mathcal{T}_i$ with timestamp $t_i$, we pair a clip $\mathcal{V}_i$ with following start and end timepoints: 
\begin{equation}
	[t_i^{start}, t_i^{end}]=[t_i-\beta_i/2\alpha,~t_i+ \beta_i/2\alpha],
\label{pairing}
\end{equation}
which represents a window centered around the timestamp $t_i$ with temporal duration equal to $\beta_i/\alpha$. $\beta_i$ is an adjustable parameter equal to the average temporal distance between pairs of consecutive narrations, i.e., ${\sum_{j=0}^{n-1}(t_{j+1}-t_j)}/{n}$. We compute $\beta_i$ on a per video basis. Conversely, $\alpha$ is a scale factor computed as the average of all $\beta_i$ across all videos in  the \dataset~($\alpha=4.9$ seconds).
Intuitively, Eq.~\ref{pairing} is derived from three observations:
\textbf{(i)}~Centering $t_i$ helps involve prior information about the event $i$;
\textbf{(ii)}~$\beta_i$ measures the clip duration according to its scenario, such as longer clips watching television~($352.9$ seconds) {v.s.} shorter clips harvesting crops~($0.9$ seconds);
\textbf{(iii)}~$\alpha$ controls the context granularity of clips (e.g.,
a large $\alpha$ pays more attention to rapid, atomic actions).
We ablate these design choices in our experimental section.
\section{Video-Language Pretraining Model}
To efficiently transfer video-language representation to egocentric downstream tasks (e.g., video-text retrieval on \epic~\cite{damen2022rescaling}), We prefer the dual-encoder~(discussed in Sec.~\ref{relatedwork_vlp}) as our VLP model architecture.
In particular, we emphasize devising a general pretraining objective \model~to adapt the existing VLP model to the egocentric domain (e.g., \dataset).

\subsection{Architecture: Dual-encoder Pipeline}
We choose Frozen~\cite{bain2021frozen} as our pretraining architecture. Frozen~\cite{bain2021frozen} design encompasses an elegant and simple dual encoder strategy (one per modality) which has favorable characteristics (e.g., indexability and efficiency~\cite{escorcia2019temporal,miech2021thinking}). Note that this allows us to use our pretrained network in single-modality tasks (e.g., video-only tasks). In practice, the video encoder adopts the TimeSformer~\cite{timesformer} architecture, while the text encoder builds upon DistillBERT~\cite{distilbert}. However, our approach is not limited to the encoder's design (e.g., the video backbone can be replaced by SlowFast~\cite{slowfast} or Video Swin~\cite{liu2022video}). 
In the rest of the paper we adopt this notation: $(\mathcal{V}_i, \mathcal{T}_i)$ represents the video-text input to the model, while $\mathbf{v}_i$ and $\mathbf{t}_i$ are used to identify the video and text embeddings.

\subsection{EgoNCE: An Egocentric-friendly Pretraining Objective}
A common pretraining objective for the dual-encoder VLP is \textbf{InfoNCE}~\cite{infonce}, where the matching visual-text pairs in the batch are treated as positives while all other pairwise combinations in the batch are regarded as negatives. 
Formally, within a batch $\mathcal{B}=\{1,\cdots, N\}$, InfoNCE is computed by the sum of the video-to-text loss $\mathcal{L}_\text{v2t}$ and text-to-video loss $\mathcal{L}_\text{t2v}$.
For simplicity, we only formulate $\mathcal{L}_\text{v2t}$, whereas $\mathcal{L}_\text{t2v}$ is defined in a symmetric way:
\begin{equation}
\mathcal{L}_\text{v2t}=\frac{1}{|\mathcal{B}|}\sum_{i\in \mathcal{B}} \log \frac{\exp(\mathbf{v}_i^T\mathbf{t}_i /\tau)}{\sum_{j\in \mathcal{B}} \exp( \mathbf{v}_i^T\mathbf{t}_j /\tau)},
\label{nce}
\end{equation}
where the $i$-th video embedding $\mathbf{v}_i$ and $j$-th text embedding $\mathbf{t}_j$ are $L_2$ normalized features, and $\tau$ is a temperature factor.

However, this simple objective performs not well on large-scale video-text datasets like \howto~\cite{miech2019howto100m} due to the serious misalignment between the two modalities of data.
Therefore, ~\cite{mil_nce} proposes MIL-NCE which treats temporal nearest captions as positive samples. 

In this work, our 1st-person human daily activity dataset, i.e. EgoClip, presents two unique challenges compared to the existing 3rd-person view video-text datasets: \textbf{Challenge (i)}: The \textcolor{Fuchsia}{same action} often occurs in  \textcolor{citecolor}{different scenarios}~(e.g.,  ``\textcolor{Fuchsia}{unlock the phone}'' could happen when ``\textcolor{citecolor}{lying in bed}'' or ``\textcolor{citecolor}{walking outdoors}'').
\textbf{Challenge (ii)}: Often, \textcolor{Fuchsia}{different actions} appearing in the \textcolor{citecolor}{same scenario} tend to have indistinguishable visual differences~(e.g.,  when ``\textcolor{citecolor}{working in front of the laptop}'', ``\textcolor{Fuchsia}{typing on the keyboard}'' or ``\textcolor{Fuchsia}{moving the mouse}'' have similar feature representations).

To overcome these two unique challenges, we propose a novel EgoNCE training objective which takes into account two simple yet efficient sampling strategies based on the vanilla InfoNCE.

\textbf{Action-aware Positive Sampling.}  
In this work, we make a reasonable assumption that the critical elements in linking visual actions to textual narrations are verbs and objects mentioned in the narrations (e.g., ``drinking coffee'' and ``opening fridge''). Following this assumption, we can devise a clever method to address challenge (i). 
Specifically, for each narration, we identify its nouns and verbs and merge synonym words based on the Ego4D taxonomy dictionary~\cite{grauman2021ego4d}, a thesaurus recording meaningful nouns/verbs in Ego4D narrations.
Then, batch samples that shared at least one noun and at least one verb are treated as positive samples.
At last, for the sample $i$, we define its positive samples set within batch 
$\mathcal{B}$ as $\mathcal{P}_i=\{j\in \mathcal{B}~|~\text{noun}(j)\cap\text{noun}(i)\neq\varnothing, \text{verb}(j)\cap\text{verb}(i)\neq\varnothing\}$.

\textbf{Scene-aware Negative Sampling.}
To address challenge (ii), we consider different actions in the same scenario as hard negative samples.
Specifically, for each video clip $i$, we sample an adjacent clip $i'\in \mathcal{N}(i)$, which is close to $i$ in time within the same video. 
We augment the original batch $\mathcal{B}$ with such hard negative samples and each sample $i$ in $\mathcal{B}$ has its negative counterparts $i'$.
Hence the batch is updated as $\mathcal{\widetilde{B}}=\{\underbrace{1,2,\cdots N}_{\mathcal{B}}, \underbrace{1',2',\cdots, N'}_{\mathcal{N}(\mathcal{B})} \}$. 

With these two sampling strategies, our new pretraining objective \textbf{\model} can be formulated as:
\begin{equation}
\begin{split}
	\mathcal{L}^\text{ego}_\text{v2t}=\frac{1}{| \mathcal{\widetilde{B}} |}\sum_{i\in\mathcal{\widetilde{B}}}  \log 
	\frac{
	\textcolor{Fuchsia}{
	\sum_{k\in \mathcal{P}_i}\exp(\mathbf{v}_i^T\mathbf{t}_k /\tau)
	}
	}
	{  \sum_{j\in \mathcal{B}} \left( \exp(\mathbf{v}_i^T\mathbf{t}_j /\tau) +
	\textcolor{citecolor}{\exp(\mathbf{v}_i^T\mathbf{t}_{j'} /\tau)}  \right) 
	}.
	\label{egonce}
\end{split}
\end{equation}
Here the item in \textcolor{Fuchsia}{purple} corresponds to our proposed action-aware positive samples and \textcolor{citecolor}{blue} corresponds to our proposed scene-aware negative samples.
\model provides a general extension to adapt the existing VLP models for  video-text pretraining datasets in the egocentric domain.
\section{\eval: A Benchmark for \vlp{} Development}

\begin{figure}[!t]
	\centering
	\includegraphics[width=1.0\linewidth]{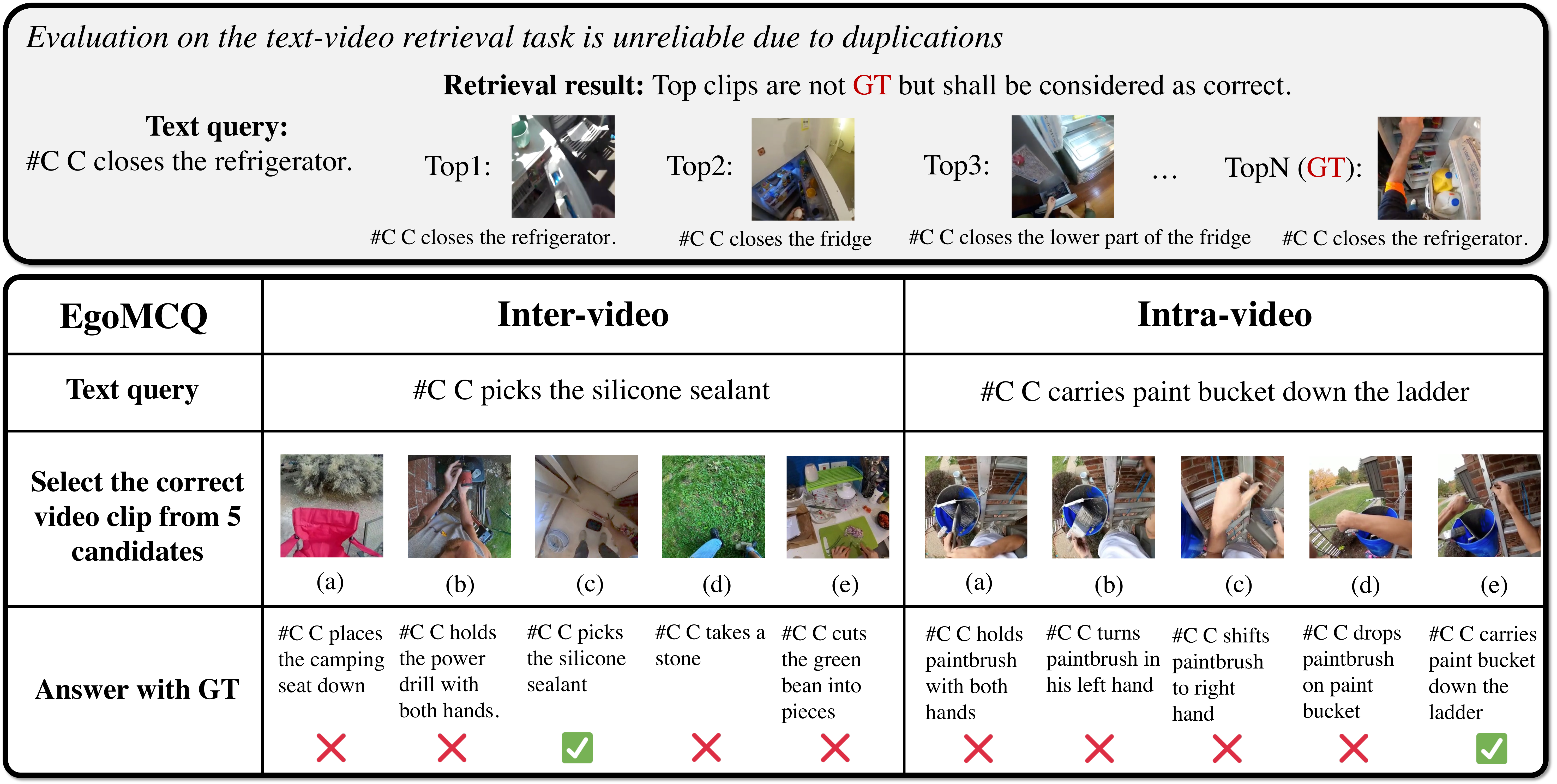}
	\vspace{-1.5em}
	\caption{
	Design of the \vlp~development set.
	\textbf{Top:} 
	An illustration of why the task of text-video retrieval is not suitable;
	\textbf{Bottom:} Two settings of EgoMCQ. 
	\textbf{Left-bottom:} The ``inter-video'' setting, each question contains $5$ clips from different videos.
	\textbf{Right-bottom:} The ``intra-video'' setting, each question contains $5$ contiguous clips from the same video, making it more challenging.} 
	\label{egomcq}
	\vspace{-1.5em}
\end{figure}

\textbf{The need for a development benchmark.}
We find that most egocentric benchmarks are domain-specific and focus on single-modality tasks (see Tab.~\ref{dataset}). However, our purpose is to exploit Ego4D's diversity to learn rich video-text representations.
Hence, to validate our design choices of the pretraining dataset (e.g., \dataset), and model (e.g., \model), it is essential to measure performance on a benchmark highly aligned with the pretraining task. 
Therefore, we propose \eval, a new egocentric benchmark for reliable and fast developments of \vlp. 

\textbf{Data source.}
We start from the Ego4D data excluded from constructing the \dataset, which mainly covers the validation set of the Ego4D challenge benchmarks.
Additionally, to assure that the scene is not visible during pretraining, we manually remove videos that share multiple views with the videos in \dataset.
To ensure diversity, we randomly select one annotator's narration for each video.
We follow the same clip pairing strategy as Eq.~\ref{pairing} to be consistent with the data format of \dataset.

\textbf{Benchmarking task design.}
To determine the task for development, we first consider video-text retrieval since it highly aligns with the VLP pretraining objective. 
However, as depicted in the top half of Fig.~\ref{egomcq}, for an action~(e.g., close the refrigerator), there are substantial duplicates or semantically similar captions in Ego4D. This can cause issues in retrieval evaluation~\cite{wray2021semantic} making model training unreliable.
A straightforward approach to prevent this is deduplication~(dedup), but it is challenging to devise a dedup criterion and perform well in the retrieval settings of a ``one-to-whole validation set''.
Therefore, we select the \textit{Multiple-Choice Questions (MCQ)} task for development since repetitions are highly unlikely given a small number of answers.

\textbf{Grouping strategies.}
To set up the MCQ task, a naive construction randomly groups five video clips to form options for a question. But we find randomly grouping is not challenging since options are highly likely to come from different videos and vary widely in content.
We redefine this basic setting as ``\textbf{inter-video}'' and ensure that the five clips originate from different videos, aiming to distinguish instances from different scenarios~(the left-bottom of Fig.~\ref{egomcq}).
Furthermore, we propose a more challenging setting ``\textbf{intra-video}'' by grouping five continuous clips together.
This setting is regarded as a specific form of video-text localization focused on fine-grained context clues, such as hand interaction~(the right-bottom of Fig.~\ref{egomcq}).
Dedup is performed within five options for each question for reliable assessment~(see Supp.~\ref{b1}) and we adopt accuracy as the \eval~metric.

\textbf{Statistics.}
We finalize $39$K questions covering $198$K narrations with $468$ hours of video, where the ``inter-video'' has $24$K questions covering $290.3$ hours of videos. And the ``intra-video'' has $15$K questions and covers $178.3$ hours of videos. The average duration among the five options is $34.2$ seconds~(More statistics of EgoMCQ are shown in Supplementary~\ref{b2}).
\section{Experiments}
We assess our \vlp~along two directions:
\textbf{(i)} We conduct an extensive analysis to explore key components of \vlp~(e.g., \dataset, \model,~and \eval);
\textbf{(ii)} we transfer our pretrained model to various downstream tasks to validate the quality of our video-text representation.

\subsection{Benchmarks and Settings}\label{e1}
We evaluate our VLP model on five egocentric benchmarks, spanning video-text tasks and pure video tasks, across three different datasets. We briefly describe each task below.

\textbf{\mir~of \epic.} 
This task is modelled as a video-text retrieval which considers the semantic overlap between different videos narrations, where multiple videos may correspond to the same narration. 
The training set contains $67.2$K clips and validation set contains $9.7$K clips. 
The evaluation metrics are mean Average Precision~(mAP) and the normalized Discounted Cumulative Gain~(nDCG).

\textbf{\nlq~of Ego4D Challenges.} 
The \nlq~task is modelled as a natural language grounding problem~\cite{hendricks2018localizing,Gao_2017_ICCV,soldan2021vlg}. 
Given a language query and a video, the task aims at localizing the temporal interval within the video, in which the answer is deducible. 
The training set contains $11.3$K queries annotated from $1$K clips for this task, while the validation contains $3.9$K queries collected from $0.3$K clips. 
The evaluation metric is Recall@$K$ for IoU${=}\theta$ (R@$K$-IoU${=}\theta$)~\cite{hendricks2018localizing} where $\theta$ is a threshold.
We evaluate for $K{\in}\{1,5\}$ and $\theta{\in}\{0.3,0.5\}$.

\textbf{Action Recognition of Charades-Ego.}
This dataset has $64$K instances, spanning 1st-person and 3rd-person views and covering $157$ activity categories for training. We train and evaluate only on the 1st-person videos. 
The validation set contains $847$ videos for classification and each video belongs to multiple classes. 
The evaluation metric is mAP. 

\textbf{\mq~of Ego4D Challenges.}
The \mq~task is a video-only task modelled as Temporal Action Localization~\cite{caba2015activitynet}. Given a particular high-level activity category, the task solution consists of retrieving all the possible temporal windows where the activity occurs. 
The training set contains $13.6$K instances from $1.5$K clips, while the validation set contains $4.3$K instances from $0.5$K clips. 
The evaluation metrics are mAP and R@$K$-IoU${=}\theta$ for $K{\in}\{1,5\}$ and $\theta{\in}\{0.3,0.5,0.7\}$.

\textbf{\ossc~(OSCC) of Ego4D Challenges.}
This OSCC task is modelled as an (N+1)-way classification aiming to identify an object's state change in a given video.
The training and val. sets contain $41$K and $28$K clips, respectively.
The evaluation metric is accuracy. 

\textbf{Implementation Details.} Our codebase is based on the official Frozen~\footnote{https://github.com/m-bain/frozen-in-time} one and retains the same settings unless specified. During pretraining, 
we sample $4$ frames for each clip,
and use the Adam optimizer~\cite{kingma2014adam} with a learning rate of $3{\times}10^{-5}$. To select the best method we pretrain our architecture for $10$ epochs and use the best performing model on the \eval~benchmark. Pretraining takes two days on $32$ A100 GPUs~($1,536$ GPU hrs).

\begin{table}[!t]
\small
\centering
\setlength{\tabcolsep}{3pt}
\vspace{-.8em}
\resizebox{1\textwidth}{!}{%
\begin{tabular}{ll|c|cc|cc}
	\toprule[1pt] 
	\multicolumn{2}{l|}{\multirow{2}{*}{\textbf{Clip creation strategy}}} & \multicolumn{1}{c|}{\textbf{Clip's length~(s)}} & \multicolumn{2}{c|}{\textbf{ EgoMCQ  Acc~(\%)}} &  \multicolumn{2}{c}{\textbf{Zero-shot T$\leftrightarrow$V Retrieval~\cite{damen2022rescaling}}} \\
	&& Avg~$\pm$~Std &  Inter-video & Intra-video & mAP~(avg) & nDCG~(avg) \\      		\midrule[1pt] 
	(a)&$[t_i,t_i{+}\alpha]$  & $5.0 \pm 0.0$                 & $87.66$ & $39.72$ & $19.6$ & $12.3$ \\ 
	(b)&$[t_i{-}\alpha/2,t_i{+}\alpha/2]$  & $5.0 \pm 0.0$    & $89.23$ & $41.68$ & $20.6$ & $13.7$ \\ 
	(c)&$[t_{i-1},t_{i+1}]$ &  $10.0 \pm 38.2 $               & $88.13$ & $40.62$ & $20.6$ & $13.7$ \\ 	 		
	(d)&$[t_i{-}\beta_i/2,t_i{+}\beta_i/2]$  &  $4.9 \pm 4.7$ & \underline{$89.74$} & $44.82$ &$21.1$ & $14.5$ \\
	(e)&$[t_i{-}\beta_i/4,t_i{+}\beta_i/4]$  &  $2.4\pm2.4$   & $\mathbf{90.23}$ & \underline{$49.67$} & \underline{$21.9$} & \underline{$15.3$} \\
	(f)&$[t_i{-}\beta_i/2\alpha,t_i{+}\beta_i/2\alpha]$  & $1.0 \pm 0.9$ & $89.36$ & $\mathbf{51.51}$ & $\mathbf{22.1}$ & $\mathbf{15.5}$ \\
	\bottomrule[1pt] 
\end{tabular}
}
\centering
\vspace{.5em}
\caption{Results on our development set \eval~and video-text retrieval on \epic~when using different strategies in the creation of \dataset, where $t_i$, $\alpha$, $\beta_i$ are defined in Eq.~\ref{pairing}.\\
In all experiments, we bold the \textbf{best results} and underlined \underline{the second best results}.
}
\vspace{-2.4em}
\label{res_egoclip}
\end{table}
\subsection{Ablation Studies}
\paragraph{Ablation of the strategy used when creating \dataset.}
We validate our proposed strategies, i.e., Eq.\ref{pairing} in Tab.~\ref{res_egoclip}, by comparing the following variants:
(a) fixed length $\alpha$, start at timestamp;
(b) fixed length $\alpha$, center at timestamp;
(c) variable clip, start and end by adjacent timestamps;
(d) our proposed strategy, scaled by $2$;
(e) our proposed strategy, scaled by $4$;
(f) our proposed strategy.

We consider that a good pretraining dataset creation strategy should satisfy: 
\textbf{(1)}~the VLP model trained on \dataset~should be able to well distinguish instances in \eval~with the same data format;
\textbf{(2)}~the VLP model pretrained on \dataset~with the specific clip creation strategy should perform well on public downstream tasks~(e.g., video-text retrieval on \cite{damen2022rescaling} and zero-shot for efficiency).

We draw several conclusions from Tab.~\ref{res_egoclip}:
\textbf{(i)}~
The performance of \eval~is well aligned with the zero-shot result on \epic, especially minor gain on downstream but noticeable on \eval,
which means \eval~provides valid feedback and is suitable as a development set.
\textbf{(ii)}~Under the same clip length $\alpha$, (b) surpassing (a) proves that centering at timestamp includes prior information is helpful.
\textbf{(iii)}~Variable-length clips make a big difference, 
as shown in (c) and (d).

\begin{wraptable}{rh}{0.45\textwidth}
\small
\centering
\vspace{-1.7em}
\setlength{\tabcolsep}{3pt}
\resizebox{0.45\textwidth}{!}{%
\begin{tabular}{l|cc}
    \toprule[1pt] 
	\multicolumn{1}{l|}{\multirow{2}{*}{\textbf{Variants}}} &  \multicolumn{2}{c}{\textbf{Accuracy~(\%)}} \\
	 & Intra-video      & Inter-video       \\ \midrule[1pt] 
	InfoNCE &     $89.4$      &     $51.5$       \\
	\midrule
	(a) w/ Pos, noun &  $82.9$~($6.5\downarrow$) & $42.3$~($9.2\downarrow$)                                 \\
	(b) w/ Pos, verb &  $86.9$~($2.5\downarrow$) & $50.5$~($1.0\downarrow$)                                 \\
	(c) w/ Pos, noun \& verb &  \underline{$89.7$}~($0.4\uparrow$) & $53.6$~($2.1\uparrow$)                 \\
	\midrule
	(d) w/ Neg, random       &  $88.3$~($1.1\downarrow$)           &  $49.9$~($1.6\downarrow$)              \\
	(e) w/ Neg, within video &  \underline{$89.7$}~($0.3\uparrow$) &  $53.0$~($1.5\uparrow$)                \\
	(f) w/ Neg, within 1 min &  {$89.5$}~($0.2\uparrow$)           & \underline{$54.5$}~($3.0\uparrow$)     \\
	\midrule
	(g)~w/~Pos~\&~Neg, \textbf{\model}   &  $\mathbf{90.6}$~($1.3\uparrow$)    &    $\mathbf{57.2}$~($5.7\uparrow$)        \\
	\bottomrule[1pt]
\end{tabular}
}
\centering
\vspace*{-8pt}
\caption{\model sampling strategy ablation. We evaluate accuracy performance on our development benchmark~\eval.}
\vspace{-1em}
\label{ablation_pos_neg}
\end{wraptable}
Notably, with our designed $\beta_i$, (d) outperforms (b) with a similar average clip length, which validates our key idea of ``contextual varied clip length''.
\textbf{(iv)}~Based on (d), (e), and (f), we found a proper scale factor greater than $1$ is preferred, which helps focus on a large of instantaneous actions densely labeled by Ego4D~\cite{grauman2021ego4d}.
These ablation studies demonstrate the effectiveness of our proposed \dataset~creation strategy and \eval~for development.

\textbf{Effect of \model.}
In this section, we evaluate the effect of the proposed sampling strategies for the \model~objective~(Eq.~\ref{egonce}) on \eval~and compare against a vanilla InfoNCE loss~(Eq.~\ref{nce}). We ablate several configurations for positive and negative sampling strategies. 
The sampling strategy for positive pairs exploits language cues, while negative pairs rely on temporal, visual cues.
Given a text-video pair, we regard other text-video pairs as positive if the textual narrations: 
(a) share at least one noun, 
(b) share at least one verb, and (c) share at least a verb-noun pair. 
Conversely, we define the following heuristics for negative sampling: 
(d) a random text-video pair from \dataset, 
(e) a text-video pair from the same video, and
(f) a text-video pair within $1$ minute from the given video-text pair annotation timestamp. 
Tab.~\ref{ablation_pos_neg} shows that using solely verbs~(a) or nouns~(b) for positive selection degrades the accuracy performance with respect to naive InfoNCE. However, we successfully push the performance beyond the baseline results when considering both verbs and nouns jointly~(c). 
Moreover, we notice that merely selecting negatives within the same video leads to better performance. In particular, we obtain the best performance for temporally ``hard negatives'' (f).
Finally, we pick the optimal settings from positive and negative sides and combine them together for (g)~\model~and reach the best results.
\subsection{Comparisons with State-of-the-arts}\label{e2}
\begin{table}[t]
\small
\centering
\vspace{-1.em}
\resizebox{1\textwidth}{!}{%
\begin{tabular}{lccc|cccccc}
    \toprule[1pt]
    \multirow{2}{*}{\textbf{Methods}} & \multirow{2}{*}{\textbf{Vis Enc Input}} &\multirow{2}{*}{\textbf{\# Frames}}  & \multirow{2}{*}{\textbf{Vis-text PT}} & \multicolumn{3}{c}{\textbf{mAP~(\%)}} & \multicolumn{3}{c}{\textbf{nDCG~(\%)}} \\
    &  & & & V$\rightarrow$T & T$\rightarrow$V & Avg & V$\rightarrow$T & T$\rightarrow$V  & Avg \\ 
    \midrule[1pt] 
    Random & - &-&-                                                             & $5.7$  & $5.6$  & $5.7$  & $10.8$ & $10.9$ & $10.9$. \\ 
    MI-MM& S3D~\cite{xie2018rethinking} & $32$ & HowTo100M                      & $34.8$ & $23.6$ & $29.2$ & $47.1$ & $42.4$ & $44.7$  \\
    MME~\cite{wray2019fine}   & TBN~$\dagger$~\cite{kazakos2019epic} & $25$ & - & $43.0$ & $34.0$ & $38.5$ & $50.1$ & $46.9$ & $48.5$  \\
    JPoSE~\cite{wray2019fine} & TBN~$\dagger$~\cite{kazakos2019epic} & $25$ & - & $49.9$ & $38.1$ & $44.0$ & $55.5$ & $51.6$ & $53.5$  \\ 
    \midrule
    Frozen~ & Raw Videos& $4$ & -             & $38.8$ & $29.7$ & $34.2$ & $50.5$ & $48.3$ & $49.4$ \\
    Frozen~ & Raw Videos& $4$ & \howto     & $39.2$ & $30.1$ & $34.7$ & $50.7$ & $48.7$ & $49.7$ \\
    Frozen~ & Raw Videos& $4$ & \ccweb & $41.2$ & $31.6$ & $36.4$ & $52.7$ & $50.2$ & $51.4$ \\
    Frozen~ & Raw Videos& $4$ &  \dataset     & \underline{$44.5$} & \underline{$34.7$} & \underline{$39.6$} & \underline{$55.7$} & \underline{$52.9$} & \underline{$54.3$} \\ 
    Frozen+\model~& Raw Videos& $4$ & \dataset & $\mathbf{45.1}$ &  $\mathbf{35.3}$ & $\mathbf{40.2}$ & $\mathbf{56.2}$ & $\mathbf{53.5}$ & $\mathbf{54.8}$\\ 
    \midrule[1pt]
    Frozen~ & Raw Videos& 16 &  \ccweb &  \underline{$45.8$} & \underline{$36.0$}  & \underline{$40.9$}  & \underline{$57.2$} & \underline{$54.3$} & \underline{$55.8$}  \\
    Frozen+\model~& Raw Videos& 16& \dataset & $\mathbf{49.9}$ &  $\mathbf{40.1}$ &$\mathbf{45.0}$ & $\mathbf{60.9}$ & $\mathbf{57.9}$ & $\mathbf{59.4}$\\
    \midrule[1pt]
    \rowcolor[gray]{0.9}
    Frozen~ & Raw Videos& $4$ & \howto      & $6.8$ & $6.3$ & $6.5$ & $11.6$ & $12.8$ & $12.2$  \\
    \rowcolor[gray]{0.9}
    Frozen~ & Raw Videos& $4$ & \ccweb & $8.6$ & $7.4$ & $8.0$ & $14.5$ & $14.6$ & $14.5$  \\
    \rowcolor[gray]{0.9}
    Frozen~& Raw Videos& 4& \dataset           & \underline{$17.9$} & \underline{$13.1$} & \underline{$15.5$} & \underline{$23.0$} & \underline{$21.2$} & \underline{$22.1$}  \\ 
    \rowcolor[gray]{0.9}
    Frozen+\model~& Raw Videos& $4$& \dataset& $\mathbf{19.4}$ & $\mathbf{13.9}$ & $\mathbf{16.6}$ &  $\mathbf{24.1}$ & $\mathbf{22.0}$ &$\mathbf{23.1}$  \\
    \bottomrule[1pt] 
\end{tabular}
}
\centering
\vspace{.5em}
\caption{
Performance of the \epic~Multi-Instance Retrieval.
Note that TBN~$\dagger$ feature~\cite{kazakos2019epic} is a combination of three modalities: RGB, Flow and Audio.
Conversely, our approach only relies on RGB input. 
The \colorbox{gray!20}{\makebox(83,4){grey highlighted rows}} correspond to \textbf{zero-shot evaluation}.
}
\vspace{-2.6em}
\label{mir}
\end{table}

\textbf{\mir.}
In Tab.~\ref{mir}, we report both zero-shot and fine-tuning evaluation results. In the zero-shot setting, 
pretraining with \dataset~($3.8$M),~despite being smaller in scale, still outperforms  \ccweb~($5.5$M)~and \howto~($136$M), validating the unique benefit of pretraining on egocentric data.
When fine-tuned with $4$ frames (rows~5-9), \dataset~pretraining  maintains a margin over the best baseline \ccweb, further verifying the viewpoint domain gap within fine-tuning.
Lastly, we increase the sample frames of our finalized model as well as the best competitor \ccweb~pretraining to $16$ (rows~10-11).
As expected, performance gains accompany the frame increase. 
We deem that notable benefits come from better temporal modeling for frequent action in the 1st-person view. 
Overall, our pretraining model outperforms the best baseline (JPoSE) by $1.0$ mAP and $5.9\%$ nDCG while requiring fewer frames and input modalities.

\begin{table}[b]
\small
\centering
\vspace{-1.em}
\setlength{\tabcolsep}{3pt}
\begin{tabular}{clc|cccc}
	\toprule
	\textbf{Methods} & \multicolumn{2}{c|}{\textbf{Video-text Pre-extrated Features}} & \multicolumn{2}{c}{\textbf{IoU=$0.3$}} & \multicolumn{2}{c}{\textbf{IoU=$0.5$}} \\
	& Vis-text Enc & Vis-text PT & R@$1$ & R@$5$ & R@$1$ & R@$5$ \\  \midrule[1pt] 
	2D-TAN~\cite{zhang2020learning} &  SlowFast+BERT& - & $5.04$ & $12.89$ & $2.02$ & $5.88$   \\
	VSLNet~\cite{zhang2020span}  &  SlowFast+BERT   & - & $5.45$ & $10.74$ & $3.12$ & $6.63$  \\  
	 \midrule
 	VSLNet~\cite{zhang2020span}  &  Frozen       & \howto        & $3.95$ & $8.72$  & $2.01$ & $4.62$ \\	
	VSLNet~\cite{zhang2020span}  &  Frozen       & \ccweb & $5.06$ & $10.30$ & $2.71$ & $6.69$ \\		
	VSLNet~\cite{zhang2020span}  &  Frozen       & \dataset      & \underline{$10.53$} & \underline{$17.94$} & \underline{$5.96$} & \underline{$11.85$}\\
	VSLNet~\cite{zhang2020span}  &  Frozen+\model& \dataset      & $\mathbf{10.84}$ & $\mathbf{18.84}$ & $\mathbf{6.81}$ & $\mathbf{13.45}$ \\	
	\bottomrule
\end{tabular}
\vspace{.5em}
\centering
\caption{Recall for several IoUs on the NLQ task's val. set.}
\vspace{-1em}
\label{nlq}
\end{table}
\textbf{\nlq.}
We report validation results on Tab.~\ref{nlq}. 
We adopt the same baselines as introduced
in~\cite{grauman2021ego4d}, namely: 2DTAN~\cite{zhang2020learning} and VSLNet~\cite{zhang2020span}, and substitute the SlowFast-BERT features with our video and language representations.
We observe a large boost in performance offered by our pretrained model on all metrics. Notably, we improve R@$1$ for IoU=$0.3$ from $5.45$ to $10.84$, despite our video branch not being pre-trained on Kinetics400.
Besides, we significantly surpass VLP pretrained on \ccweb~and~\howto.
We believe that this increase is due to the egocentric data availability and the video-text interaction learned from large-scale pretraining. 
Please see Supplementary \ref{d_nlq} for the test set results.

\begin{table}[t]
\small
\centering
\resizebox{1\textwidth}{!}{%
\begin{tabular}{lcccl|c}
	\toprule[1pt] 
	\textbf{Methods} &  \textbf{Vis Enc} & \textbf{\# Frames} &  \textbf{Vis-Text PT} & \textbf{Train~/~FT Data} & \textbf{mAP~(\%)}  \\ \midrule[1pt] 
	Actor~\cite{sigurdsson2018actor} & ResNet-152  & $25$ & - & Charades-Ego~(1st + 3rd) & $20.0$  \\
	SSDA~\cite{choi2020unsupervised} & I3D         & $32$ & - & Charades-Ego~(1st + 3rd) & $23.1$ \\
	I3D~\cite{choi2020unsupervised}  & I3D         & $32$ & - & Charades-Ego~(1st).      & $25.8$  \\
	Ego-Exo~\cite{li2021ego} &  SlowFast~(ResNet-101) & $32$ & - & Charades-Ego~(1st)       & $30.1$  \\ 
	\midrule[1pt] 
	Frozen & TimeSformer  & $16$ &  -             & Charades-Ego~(1st) & $28.8$ \\
	Frozen & TimeSformer  & $16$ &  \howto     & Charades-Ego~(1st) & $28.3$ \\
	Frozen & TimeSformer  & $16$ &  \ccweb & Charades-Ego~(1st) & $30.9$ \\
	Frozen & TimeSformer  & $16$ &  \dataset      & Charades-Ego~(1st) & \underline{$31.2$}\\
	Frozen+\model&TimeSformer&$16$& \dataset      & Charades-Ego~(1st) & $\mathbf{32.1}$ \\
	\midrule[1pt] 
	\rowcolor[gray]{0.9}
	Frozen~ & TimeSformer & $16$ &  \howto     & - & $9.2$              \\ \rowcolor[gray]{0.9}
	Frozen~ & TimeSformer & $16$ &  \ccweb & - & $20.9$             \\ \rowcolor[gray]{0.9}
	Frozen~ & TimeSformer & $16$ &  \dataset      & - & \underline{$23.6$} \\ \rowcolor[gray]{0.9}
	Frozen+\model & TimeSformer & $16$ & \dataset & - & $\mathbf{25.0}$    \\
	\bottomrule[1pt] 
\end{tabular}
}
\centering
\vspace{.5em}
\caption{Performance of the action recognition on the Charades-Ego dataset (a first-person test set).
The \colorbox{gray!20}{\makebox(83,4){grey highlighted rows}} correspond to \textbf{zero-shot evaluation}.
}
\vspace{-1.5em}
\label{charades}
\end{table}
\noindent \textbf{Action Recognition.}
We conduct action recognition on Charades-Ego, where categories are short phrases like ``Holding some clothes''. Thus this task can be solved as a video-text retrieval by leveraging the text representation. 
We present the result in Tab.~\ref{charades} under zero-shot and fine-tuning settings.
In zero-shot settings,
our model outperforms two supervised baselines, which validates the stronger generalization of jointly learning video-text features.
After fine-tuning (rows~5-9), our model surpasses all VLP counterparts and improves over the state-of-the-art classifier Ego-Exo by $2.0$\% with fewer sampled frames, which shows the superior advantage of joint video-text representations.

\begin{table}[h]
\centering
\setlength{\tabcolsep}{3pt}
\resizebox{\linewidth}{!}{
\begin{tabular}{clc|cccccc|cccccc}
		\toprule[1pt] 
		\textbf{Methods} & \multicolumn{2}{c|}{\textbf{Video Pre-extracted Features}} & \multicolumn{2}{c}{\textbf{IoU=$0.3$}} & \multicolumn{2}{c}{\textbf{IoU=$0.5$}} &
		\multicolumn{2}{c|}{\textbf{IoU=$0.7$}} &
		\multicolumn{4}{c}{\textbf{mAP~(\%) @ IoU}}\\
		& Vis Enc & Vis-text PT & R@$1$ & R@$5$ &R@$1$ & R@$5$ & R@$1$ & R@$5$ & $0.1$ & $0.3$ & $0.5$ & Avg \\ 
		\midrule[1pt] 
		VSGN~\cite{zhao2021video} & SlowFast    &  -        & $33.45$ & $58.43$ & $25.16$ & $46.18$ & $15.36$ & $25.81$ & $9.10$ & $5.76$ & $3.41$ & $6.03$  \\\midrule[1pt] 
		VSGN~\cite{zhao2021video} & Frozen & \howto         & $31.40$ & $52.61$ & $22.28$ & $41.29$ & $13.41$ & $23.21$ & $9.83$ & $6.72$ & $3.84$ & $6.72$  \\		
		VSGN~\cite{zhao2021video} & Frozen  & \ccweb  & $32.08$ & $56.40$ & $23.46$ & $43.81$ & $13.73$ & $23.77$ & $9.83$ & $6.40$ & $3.86$ & $6.58$  \\		
		VSGN~\cite{zhao2021video} & Frozen & \dataset& \underline{$40.06$} & \underline{$63.71$} & \underline{$29.59$} & \underline{$48.32$} & \underline{$17.41$} & \underline{$26.33$} & \underline{$15.90$} & \underline{$10.54$} & \underline{$6.19$} & \underline{$10.69$} \\	
		VSGN~\cite{zhao2021video} & Frozen+\model~ & \dataset& $\mathbf{40.43}$ & $\mathbf{65.67}$ & $\mathbf{30.14}$ & $\mathbf{51.98}$ & $\mathbf{19.06}$ & $\mathbf{29.77}$ &  $\mathbf{16.63}$ & $\mathbf{11.45}$ & $\mathbf{6.57}$ &$\mathbf{11.39}$ \\
		\bottomrule[1pt] 
	\end{tabular}
}
\vspace{.3em}
\caption{Recall and mAP metrics for several IoUs on the Moment Query task's val. set.}
\label{mq}
\vspace{-1.0em}
\end{table}

\textbf{\mq.} 
This task investigates the quality of video-only features. We extract video features and provide them as input to the VSGN model~\cite{zhao2021video}. We report the validation results in Tab.~\ref{mq},
We find that our features achieves the best performance over SlowFast features with an increase of $4.66\%$ in Avg mAP.
Moreover, we maintain better performance with respect to 3rd-person large-scale pretraining datasets.
This demonstrates that the 1st-person VLP model also learns competitive video representations. Please see the Supplementary \ref{d_mq} for the test set results.

\begin{wraptable}{rh}{0.45\textwidth}
\small
\centering
\vspace{-1.1em}
\setlength{\tabcolsep}{3pt}
\resizebox{0.45\textwidth}{!}{%
\begin{tabular}{lc|c}
	\toprule[1pt] 
	\textbf{Methods}  &  \textbf{Vis-Text PT} & \textbf{Acc.~(\%)} \\ \midrule[1pt] 
	Always Positive                          &  -        & $48.1$  \\
	Bi-d LSTM~\cite{graves2005bidirectional} &  ImageNet & $65.3$  \\
	I3D (ResNet-50)~\cite{carreira2017quo}     &  -        & $68.7$  \\ 
	\midrule
	Frozen &   -            & $70.3$              \\
	Frozen &  \howto     & $71.7$              \\
	Frozen &  \ccweb & $71.5$              \\
	Frozen &  \dataset      & \underline{$73.4$}  \\
	Frozen+\model~ & \dataset  & $\mathbf{73.9}$ \\
	\bottomrule[1pt] 
\end{tabular}
}
\centering
\caption{Accuracy metric on the \ossc~task's val. set.}
\vspace{-2em}
\label{ossc}
\end{wraptable}\textbf{\ossc.} We report the validation results on Tab.~\ref{ossc}. Once again, our model achieves the best performance of all baselines, $2.4\%$ than \ccweb~counterparts, which indicates our visual representations are able to focus on the fine-grained clues related to state changes.

\textbf{Summary of \model.} From the above experimental results, Frozen pretrained on \dataset~with the \model objective brings a consistent improvement over the InfoNCE on all downstream tasks, which comprehensively demonstrates the effect of \model, as well as the decision  from \eval.

\newpage
\section{Conclusion, Limitations, and Societal Impacts.}
To the best of our knowledge, this work is the pioneering work to unlock \vlp.
\textbf{(i)}~We devise a principled data curation and create EgoClip, an egocentric large-scale text-video pretraining dataset with $3.8$M clip-text pairs well-chosen from Ego4D.
\textbf{(ii)}~We exploit the particular characteristics of egocentric videos and devise \model~with meaningful sampling strategies for effective egocentric pretraining.
\textbf{(iii)}~We create EgoMCQ, an egocentric video-language benchmark close to the pretraining set to support efficient exploration and development of \dataset~and \model.
Finally, we further demonstrate the strong representation of our egocentric pretraining on five tasks across three datasets. 
We believe that our EgoClip, EgoMCQ and EgoNCE would greatly benefit the egocentric video community, laying a good foundation for the new research trend of egocentric VLP. \\
\textbf{Limitations.} 
Our pretraining approach does not take into account the long-term temporal dependencies in long Ego4D videos. We leave this for future work. \\
\textbf{Societal impact.} 
\vlp~learns real-world perception knowledge that may contribute to practical applications such as augmented reality and robotics.
However, Ego4D videos collected by participants may contain users' privacy and unintended biases, so should be used cautiously. We refer the readers to the Ego4D paper about further privacy and societal impacts.
\section{Acknowledgements}
This project is supported by the National Research Foundation, Singapore under its NRFF Award NRF-NRFF13-2021-0008, and Mike Zheng Shou's Start-Up Grant from NUS. The computational work for this article was partially performed on resources of the National Supercomputing Centre, Singapore. Michael Wray and Dima Damen are supported by EPSRC UMPIRE (EP/T004991/1). Mattia Soldan and Bernard Ghanem 
are supported by the King Abdullah University of Science and Technology (KAUST) Office of Sponsored Research through the Visual Computing Center (VCC) funding, as well as, the SDAIA-KAUST Center of Excellence in Data Science and Artificial Intelligence (SDAIA-KAUST AI).
Thanks to Tencent Data Platform for the support of computing resources.
Our work is built upon the Ego4D dataset, and we greatly appreciate the contributions and efforts of the Ego4D community.

\bibliographystyle{unsrt}
\bibliography{ref}

\newcommand{\secf}{Differentiating Egocentric VLP and Ego4D}
\newcommand{\seca}{Construction details and statistics of \dataset~pretraining dataset}
\newcommand{\secb}{Construction details and statistics of \eval~benchmark}
\newcommand{\secc}{Technical details of our VLP model}
\newcommand{\secd}{Additional experimental details and results}

\newpage
\section*{Appendix}
We present the following items in the supplemental material:

\begin{enumerate}[label=\textbf{\Alph*}.]
    \item \secf~in Sec.~\ref{f}.
    \item \seca~in Sec.~\ref{a}.
    \item \secb~in Sec.~\ref{b}.
    \item \secc~in Sec.~\ref{c}.
    \item \secd~in Sec.~\ref{d}.
\end{enumerate}

\renewcommand\thesection{\Alph{section}}
\setcounter{section}{0}
\section{\secf}\label{f}
In our work, we study the video-language pretraining in a specific yet significant domain - the 1st-person view, which is motivated by the release of the Ego4D dataset. However, there is a long way to pave from the Ego4D dataset to Egocentric VLP, which consists of the pretraining dataset, development set, model designs, and transferability evaluation. Since they are not as fully explored as their third-person counterparts, thus we pioneer them by ourselves and conduct \textit{a systematic study toward the egocentric video-language pretraining} - the contribution of our work.

\subsection{Pretraining dataset}
Despite the merits of Ego4D, it has not been proposed for video-language pretraining, and cannot be directly used as its untrimmed videos, no direct video-text pairs, and noisy data. We thus see our clear distinction and contribution in proposing a successful approach to curate a pretraining dataset, our proposed EgoClip. Notably, It is also non-trivial to figure out what is the best way of curating Ego4D to create a pretraining dataset EgoClip, e.g., our pairing approach outperforms the naive strategy with a large margin in the development set, which requires substantial design and experimental validations. We add a Tab.~\ref{dataset_supp}, as an extension of Tab.~\ref{dataset}, to clearly show their difference.

\begin{table}[h]
\small
\centering
\resizebox{1.0\textwidth}{!}{%
\begin{tabular}{lccccc}
	\toprule[1pt] 
	\textbf{Dataset} & \textbf{Ego?}    & \textbf{Domain} &  \textbf{Dur~(hrs)}   &   \textbf{\#~Clips} & \textbf{\#~Texts} \\ 
	\midrule
	{Ego4D~\cite{grauman2021ego4d}~(untrimmed)}                           & {\Checkmark} & {\textbf{diverse}} & {$\mathbf{3.6K}$} & {-} & {$\mathbf{5.0M}$}  \\
	{\textbf{\dataset}~(well-curated from Ego4D)}                           & {\Checkmark} & {\textbf{diverse}} & {${2.9K}$} & {$\mathbf{3.8M}$} & {${3.8M}$} \\
	\bottomrule[1pt] 
\end{tabular}
}
\centering
\vspace{.5em}
\caption{Comparison of \dataset~and Ego4D dataset.}
\label{dataset_supp}
\vspace{-2.0em}
\end{table}

\subsection{Development set}
In the 1st-person domain, there is lacking a satisfactory benchmark that good aligns with pretraining data diversity and focuses on video-text alignment. Therefore, we propose a new development set i.e. EgoMCQ to power rapid design of video-text pretraining i.e. its pretraining dataset and model pretraining objective.

\subsection{Model designs}
We select Frozen as the baseline because its elegant and scalable dual-encoder architecture is representative in state-of-the-art VLP methods. Besides, corresponding to MIL-NCE~\cite{miech2020end} built on top of the 3rd-person domain's \howto~\cite{miech2019howto100m}, we aim to explore a general pretraining objective i.e., \model to learn rich video-text representations in 1st-person domains.

\subsection{Transferability evaluation}
Extensive experiments and promising results demonstrate the effectiveness and necessity of Egocentric VLP, which will greatly benefit the egocentric community. Note that Ego4D has not been used previously for any downstream tasks on other datasets. This is also where our work makes significant value.
\newpage
\section{\seca}\label{a}
\subsection{Data filtering}\label{a1}
After we source video-text data for \dataset, we adopt the following criteria to further reduce noise:

\textbf{(i)}~We select double-sized stereo videos~($1.3\%$ videos dur) and keep half per video for a normal size. 

\textbf{(ii)}~We discard videos with an aspect ratio greater than $2$~($0.4\%$ videos dur).

\textbf{(iii)}~We filter narrations with unsure tags~($4.0\%$ texts) e.g. ``\texttt{\#C C washes \#unsure in sink}''.

\textbf{(iv)}~We remove narrations less than $3$ words~($0.9\%$ texts),
since such narrations generally cannot be deduced from the video, 
e.g., ``\texttt{\#C C speaks}'', ``\texttt{\#C C looks}''.

\subsection{Data compression}\label{a2}
The Ego4D videos are untrimmed, 
which tend to be very long~(average $24$ mins and max to $7$ hrs) and have large resolution~(e.g., $1920\times 1080$, $1440\times 1080$),
so it is impossible to adopt untrimmed videos as model input due to heavy data loading. 
Therefore we propose to compress them:

\textbf{(i)} We first resize all videos with short size $256$.

\textbf{(ii)} Chunk each resized video into several segments, which are up to $10$ min in length.

During pretraining, given the start and end time points of a clip, we only load the segment that this clip belongs to, rather than the whole video. 
To this end, we are able to perform efficient end-to-end pretraining with raw RGB videos as model input. 
One epoch of pretraining $3.8\text{M}$ video-text pairs costs $6$ hrs on $32$ V100 GPUs~($192$ GPU hrs).

\subsection{Data analysis}\label{a3}
\begin{figure}[htb]
\vspace{-0.25cm}
    \centering
    \includegraphics[width=0.45\linewidth]{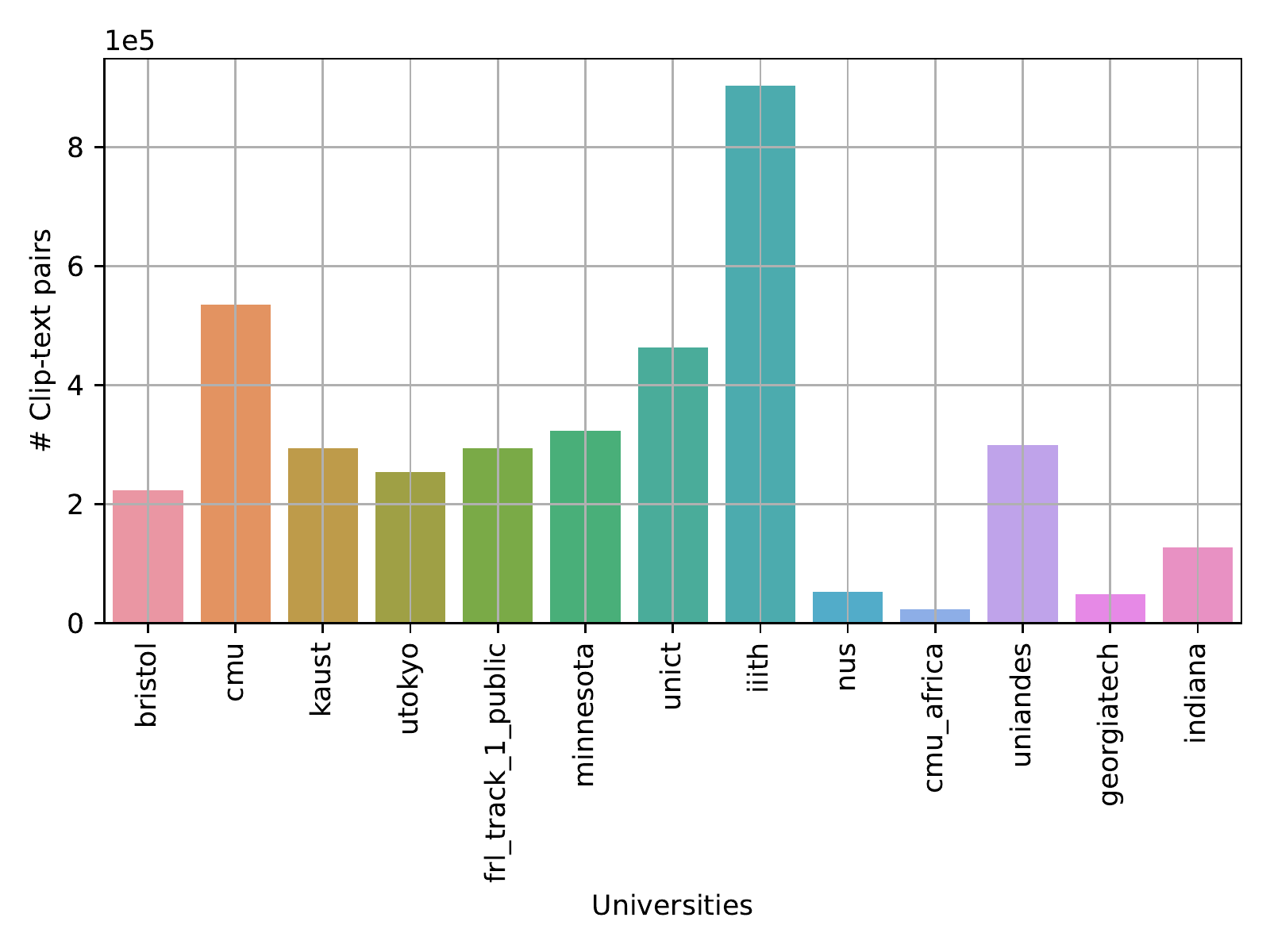}
    \caption{Institution distribution of \dataset}
    \label{fig_egoclip_country}
\vspace{-0.5cm}
\end{figure}

\noindent\textbf{Geographic diversity.}
We present the distribution of \dataset~clips source~in Fig.~\ref{fig_egoclip_country}, which covers worldwide 13 institutions from 9 different countries~\cite{grauman2021ego4d}, including:
Europe~(UK, Italy);
Asia~(India, Japan, Singapore, Kingdom of Saudi Arabia);
America~(USA, Colombia);
Africa~(Rwanda).
Therefore, our created pretraining dataset inherited the good geographic as well as participants diversities of Ego4D~
(More details can be found in ``Supp. C. Demographics'' in Ego4D paper~\cite{grauman2021ego4d}).
\newpage

\begin{figure}[H]
\centering
\begin{minipage}[t]{0.45\textwidth}
\centering
\includegraphics[width=1.0\linewidth]{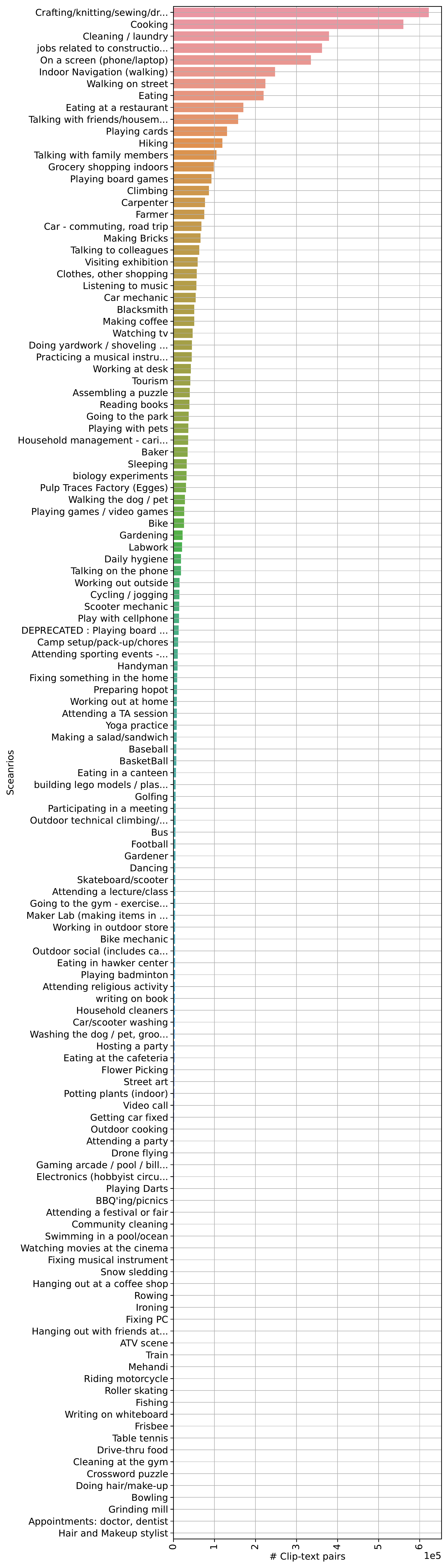}
\caption{Scenario distribution of \dataset}
\label{fig_egoclip_sceanrio}
\end{minipage}
\quad
\begin{minipage}[t]{0.45\textwidth}
\centering
\includegraphics[width=1.0\linewidth]{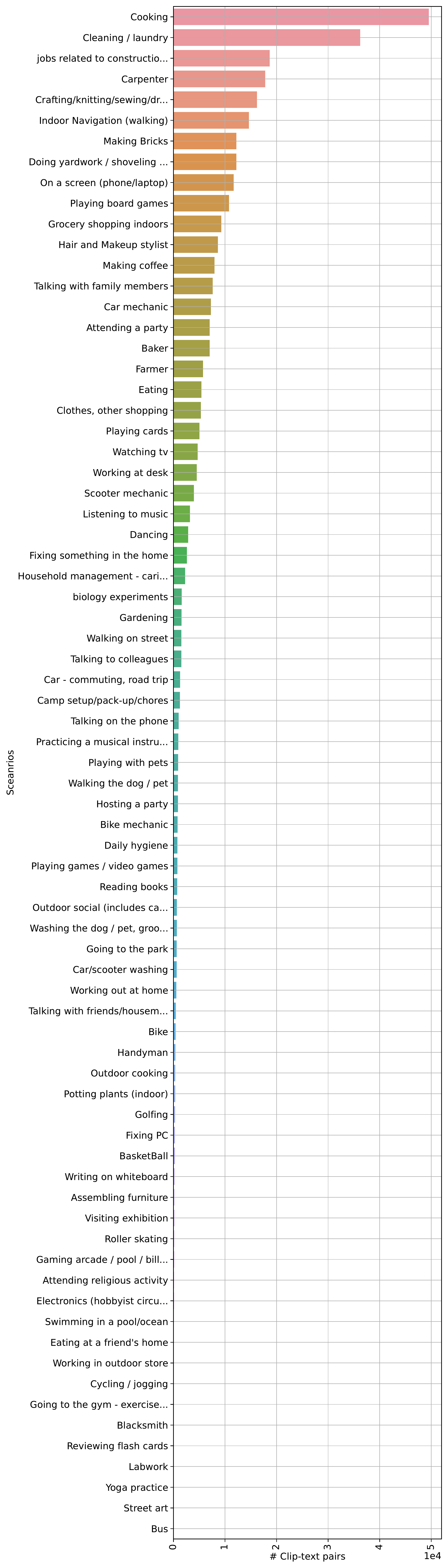}
\caption{Scenario distribution of \eval}
\label{case_1}
\end{minipage}
\centering
\label{fig_egomcq_sceanrio}
\end{figure}

\noindent\textbf{Scenario diversity.}
We have statistics the scenario distribution of \dataset~in Fig.~\ref{fig_egoclip_sceanrio}, which covers $129$ human daily scenarios e.g., household~(cooking, cleaning), outdoor~(shopping, hiking), workplace~(at desk, on a laptop), leisure~(playing board games), etc.
Notably, this distribution is long tailed, where the largest scenario ``Crafting/knitting/sewing/drawing/painting'' includes $622\text{K}~(11.1\%)$ and the smallest scenario ``Hair and Makeup stylist'' contains $35$ instances.

\begin{figure}[H]
\vspace{-0.5cm}
\centering
\subfigure[Frequency of clips]{
\begin{minipage}[t]{0.3\textwidth}
\centering
\includegraphics[width=1.0\linewidth]{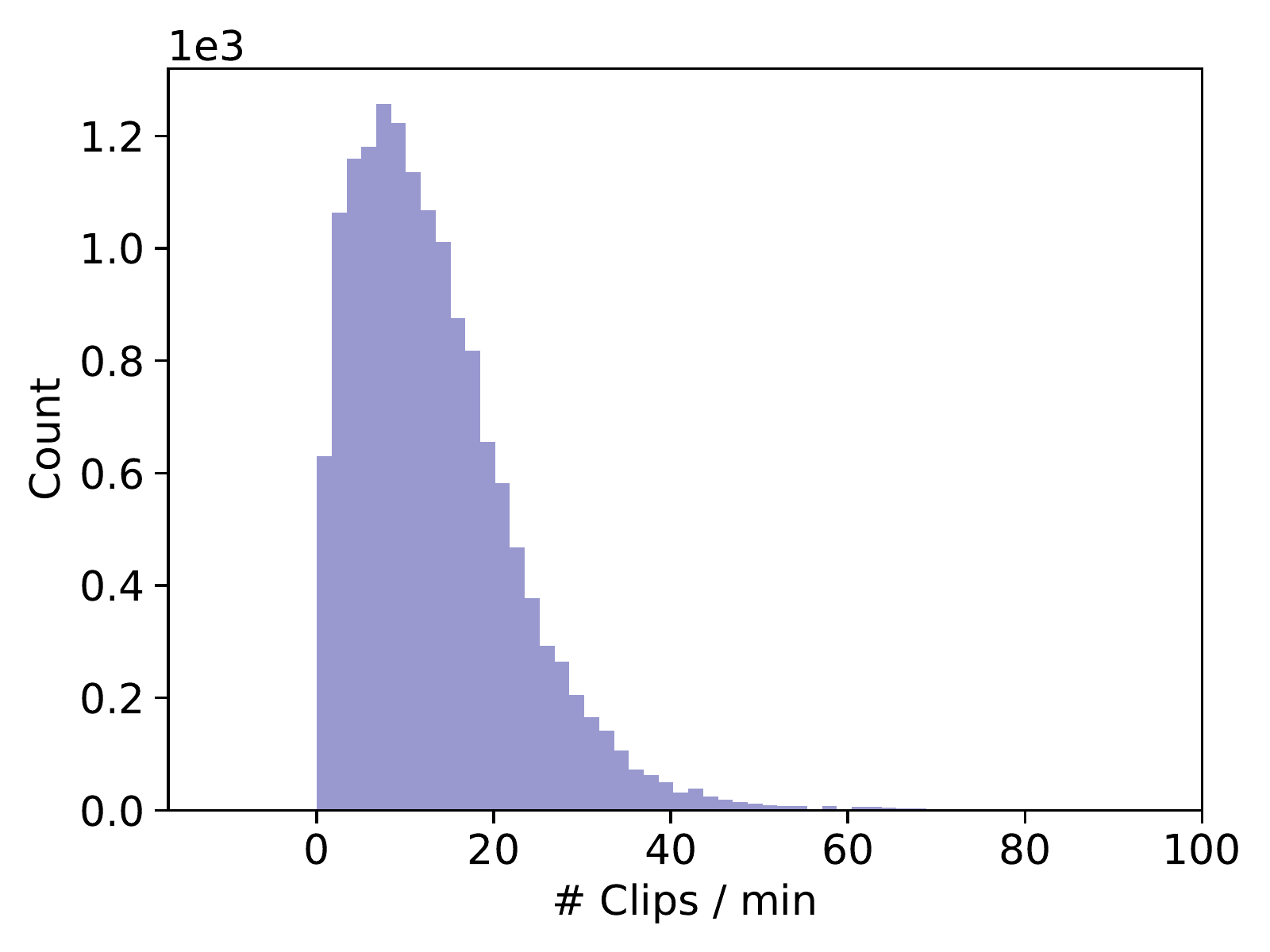}
\label{case_1}
\end{minipage}
}
\subfigure[Duration of clips]{
\begin{minipage}[t]{0.3\textwidth}
\centering
\includegraphics[width=1.0\linewidth]{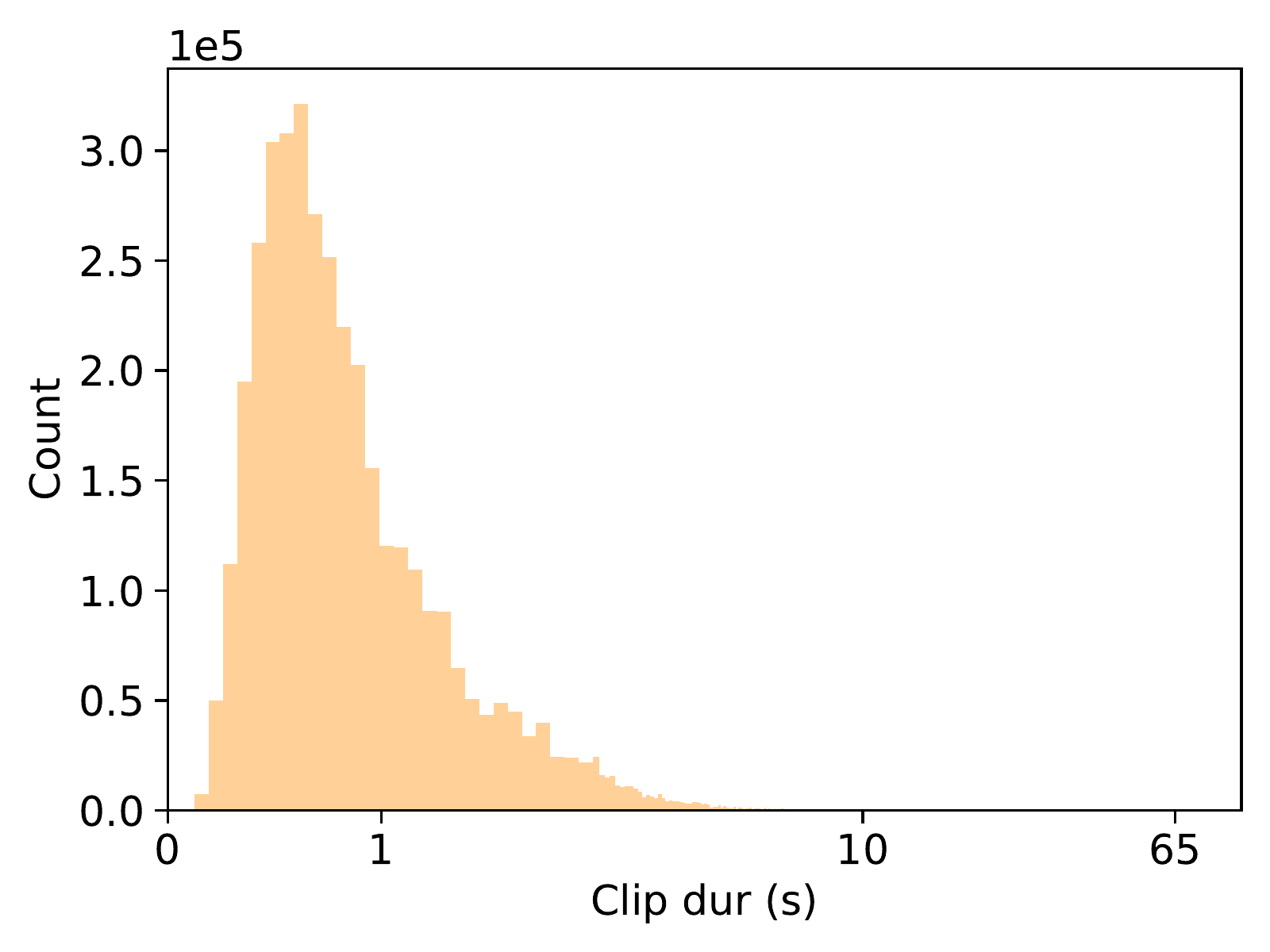}
\label{case_1}
\end{minipage}
}
\subfigure[Words length of narration]{
\begin{minipage}[t]{0.3\textwidth}
\centering
\includegraphics[width=1.0\linewidth]{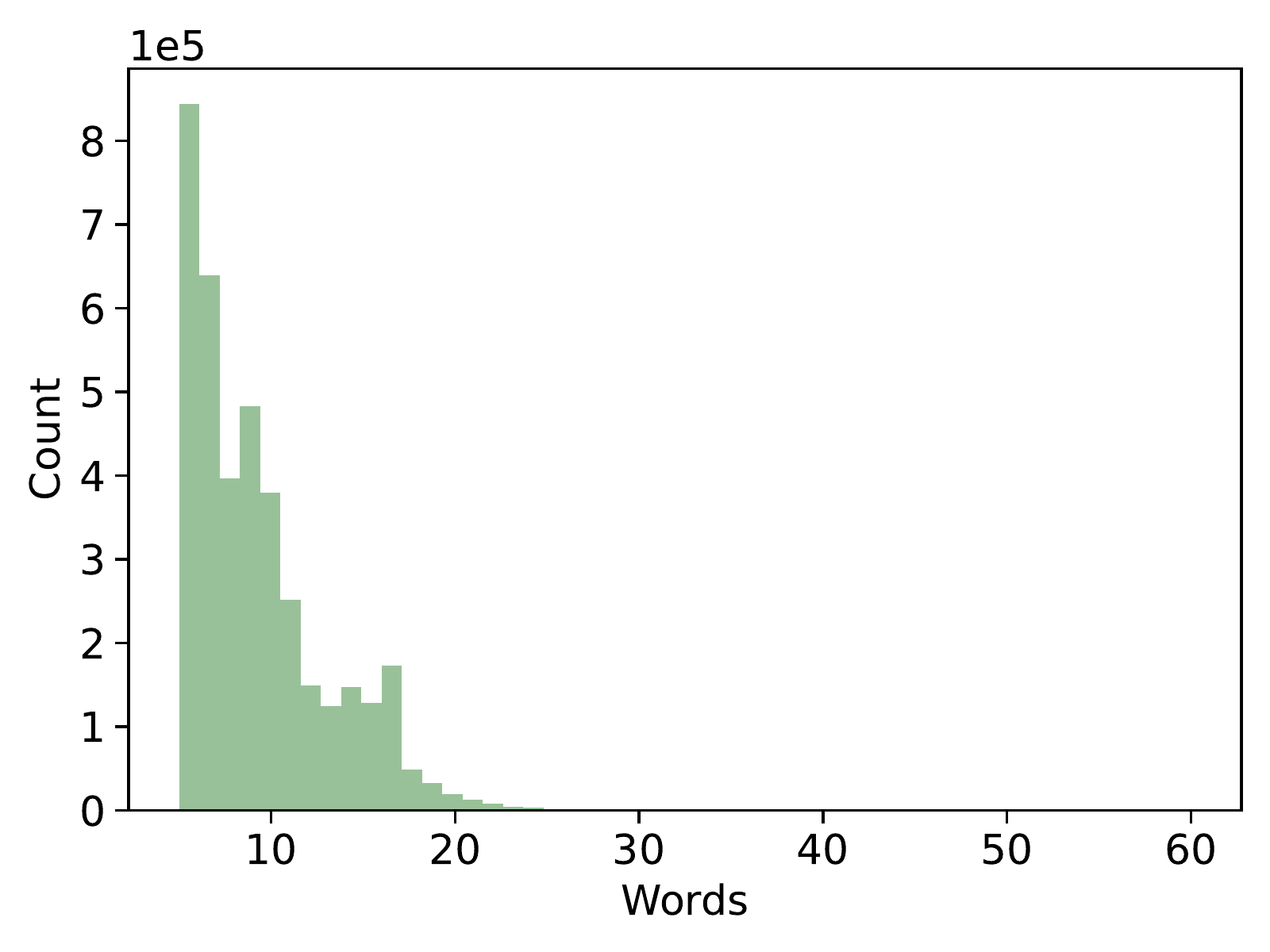}
\label{case_1}
\end{minipage}
}
\centering
\caption{Clip and narration distribution of \dataset}
\label{fig_egoclip_clip}
\vspace{-0.5cm}
\end{figure}

\noindent\textbf{Clip analysis.}
We present the statistics on the created clips in \dataset.
Fig.~\ref{fig_egoclip_clip}~(a) shows the distribution of clip frequency over the $2.9\text{K}$ pretraining set videos~(For each video, we calculate two frequencies from two annotators respectively).
The varying clip frequencies are mainly dependent on manual narrations that are annotated based on the video scenarios and activities. 
There have average $13.4$ clips per minute of video, maximize to $175.8$ narrations~/~minute and minimize to $0.06$ narrations~/~minute.
Our clip creation strategy Eq.~(1) takes this characteristic into account by estimating clip length based on the frequency of the video that the clip belongs.
Fig.~\ref{fig_egoclip_clip}~(b) displays the distribution of clip duration. 
The average duration is $0.98$ seconds with a standard deviation of $0.95$ seconds, and $69.5\%$ of clips are less than $1.0$ seconds in length, 
due to the massive atomic instantaneous actions densely labeled by Ego4D. 
Besides, the clip might be max to $65.36$ seconds, which corresponding to the scenario that ``a people walking in a forest''.

\noindent\textbf{Narration analysis.}
In Fig.~\ref{fig_egoclip_clip}~(c), we present the distribution of narration words length.
The average words length of \dataset~narration is $9.39$.
Notably, the \dataset~narrations cover $116$ verbs and $555$ nouns, where we merge the semantically synonyms words, e.g., the nouns of ``handkerchief'',``napkin'',``serviette'',``tissue'',``wipe'' both belong to ``napkin''.
Each narration of \dataset~have $1.84$ nouns and $0.87$ verbs on average.

\begin{figure}[htb]
\vspace{-0.25cm}
\centering
\subfigure[Top 50 most frequently verbs distribution]{
\begin{minipage}[t]{0.9\textwidth}
\centering
\includegraphics[width=1.0\linewidth]{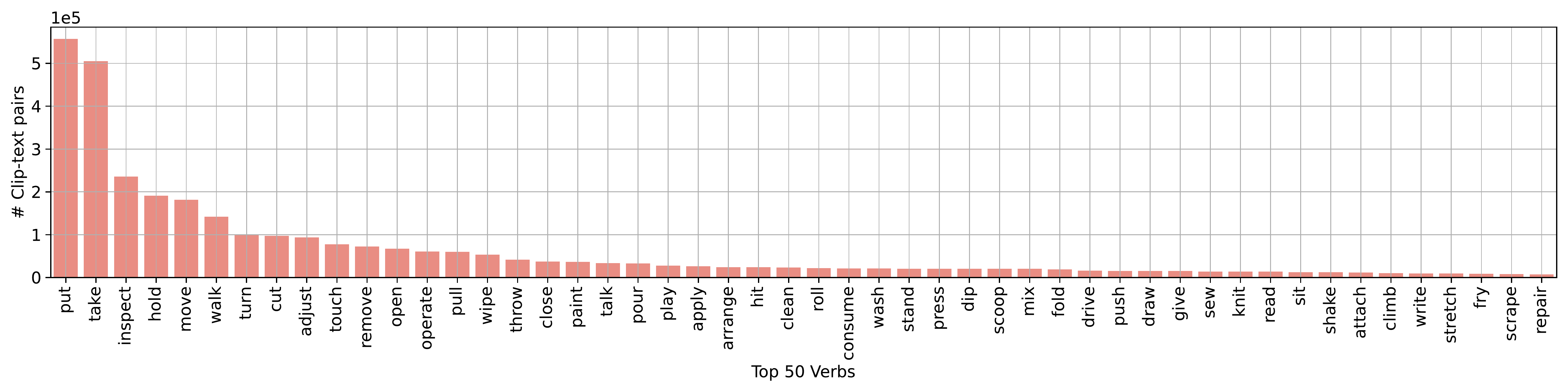}
\label{case_1}
\end{minipage}
}
\vspace{-0.25cm}
\subfigure[Top 50 most frequently nouns distribution]{
\begin{minipage}[t]{0.9\textwidth}
\centering
\includegraphics[width=1.0\linewidth]{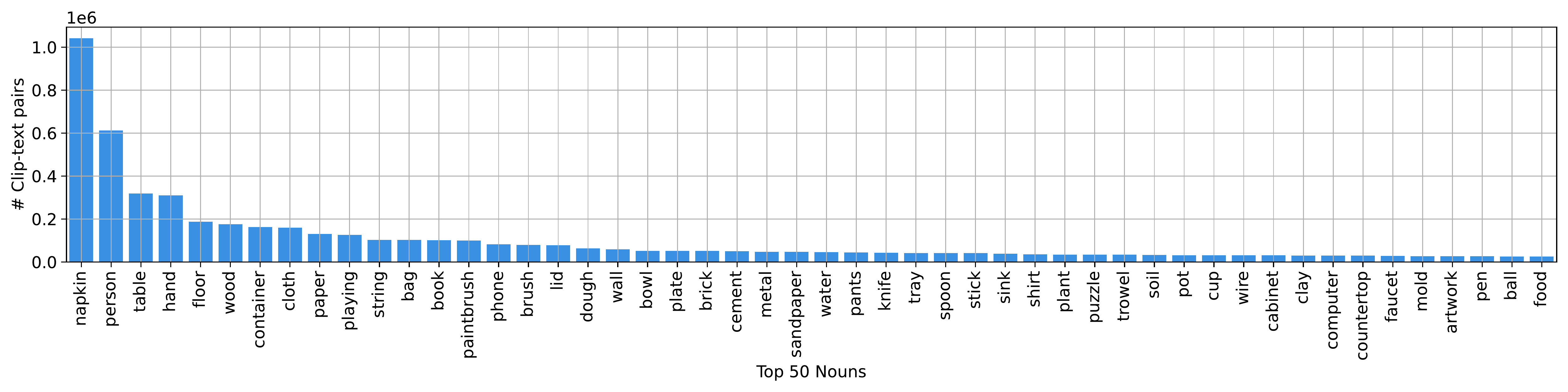}
\label{case_1}
\end{minipage}
}
\centering
\caption{Verbs and nouns distributions of \dataset’s narrations}
\label{fig_egoclip_v_n}
\vspace{-0.25cm}
\end{figure}
We further display the distribution of the top 50 most frequently verbs and nouns of \dataset~in Fig.~\ref{fig_egoclip_v_n}.
The most common nouns is ``napkin'', which appeared in $1.0\text{M}~(27.06\%)$ clips. 

\begin{figure}[H]
    \vspace{-0.25cm}
    \centering
    \subfigure[\texttt{\#C C ties the vegetable with a band.}]{
    \begin{minipage}[]{0.9\textwidth}
    \includegraphics[width=0.19\textwidth]{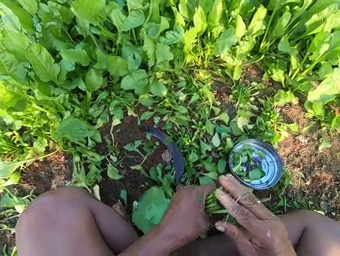}\vspace{4pt}
    \includegraphics[width=0.19\textwidth]{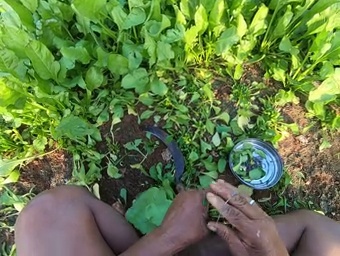}\vspace{4pt}
    \includegraphics[width=0.19\textwidth]{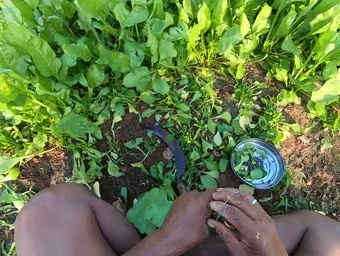}\vspace{4pt}
    \includegraphics[width=0.19\textwidth]{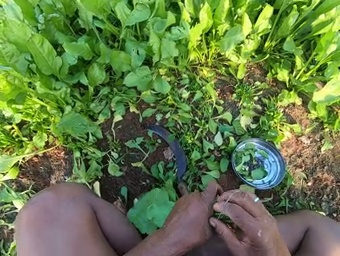}\vspace{4pt}
    \includegraphics[width=0.19\textwidth]{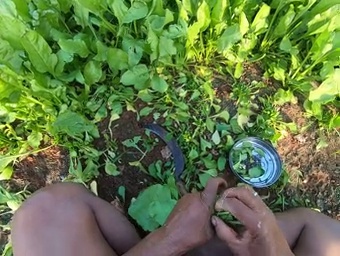}
    \end{minipage}
    }
    \subfigure[\texttt{\#C C moves the right hand.}]{
    \begin{minipage}[]{0.9\textwidth}
    \includegraphics[width=0.19\textwidth]{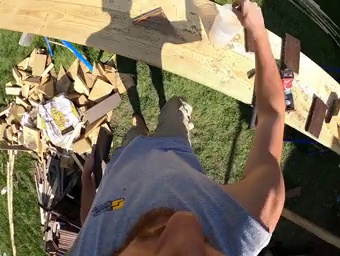}\vspace{4pt}
    \includegraphics[width=0.19\textwidth]{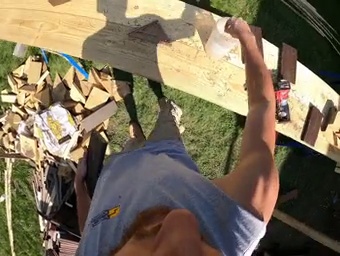}\vspace{4pt}
    \includegraphics[width=0.19\textwidth]{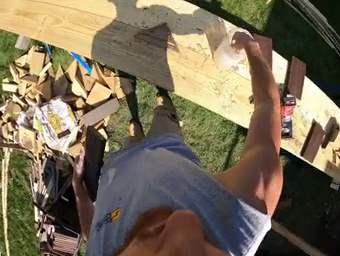}\vspace{4pt}
    \includegraphics[width=0.19\textwidth]{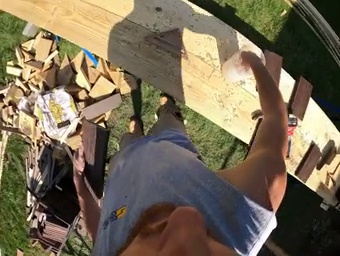}\vspace{4pt}
    \includegraphics[width=0.19\textwidth]{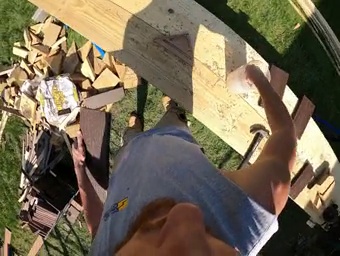}
    \end{minipage}
    }
    \subfigure[\texttt{\#C C picks the chopsticks.}]{
    \begin{minipage}[]{0.9\textwidth}
    \includegraphics[width=0.19\textwidth]{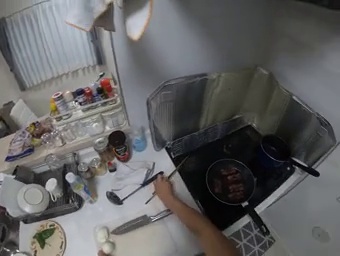}\vspace{4pt}
    \includegraphics[width=0.19\textwidth]{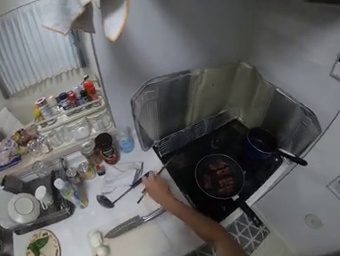}\vspace{4pt}
    \includegraphics[width=0.19\textwidth]{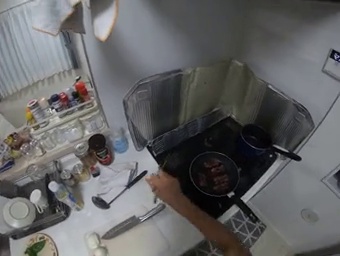}\vspace{4pt}
    \includegraphics[width=0.19\textwidth]{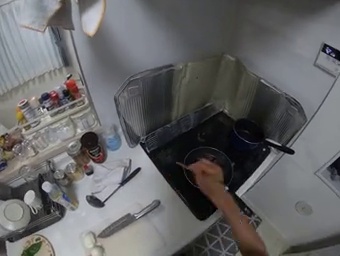}\vspace{4pt}
    \includegraphics[width=0.19\textwidth]{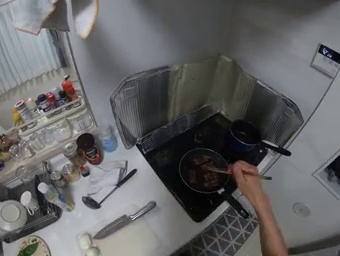}
    \end{minipage}
    }
    \subfigure[\texttt{\#C C cuts the apple with a knife.}]{
    \begin{minipage}[]{0.9\textwidth}
    \includegraphics[width=0.19\textwidth]{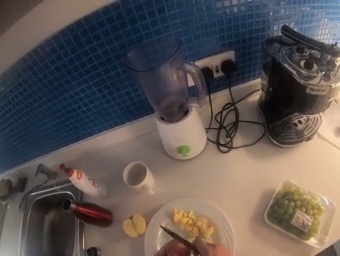}\vspace{4pt}
    \includegraphics[width=0.19\textwidth]{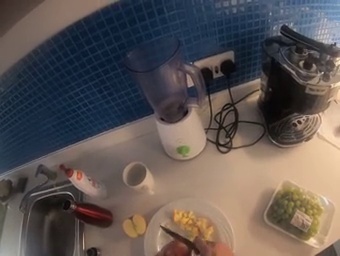}\vspace{4pt}
    \includegraphics[width=0.19\textwidth]{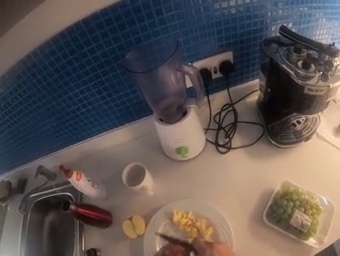}\vspace{4pt}
    \includegraphics[width=0.19\textwidth]{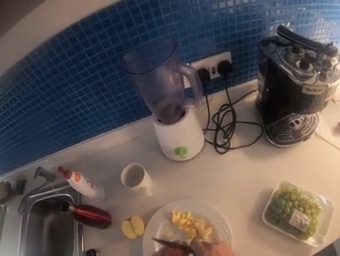}\vspace{4pt}
    \includegraphics[width=0.19\textwidth]{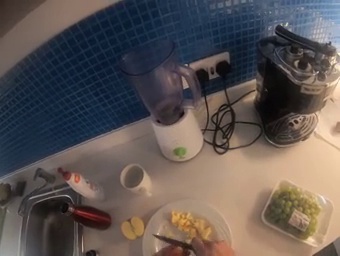}
    \end{minipage}
    }
    \subfigure[\texttt{\#C C draws on a book.}]{
    \begin{minipage}[]{0.9\textwidth}
    \includegraphics[width=0.19\textwidth]{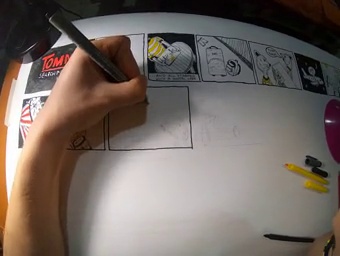}\vspace{4pt}
    \includegraphics[width=0.19\textwidth]{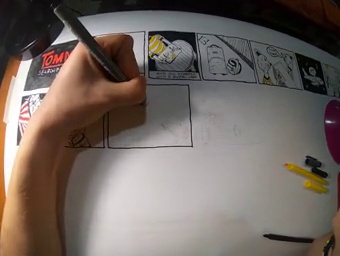}\vspace{4pt}
    \includegraphics[width=0.19\textwidth]{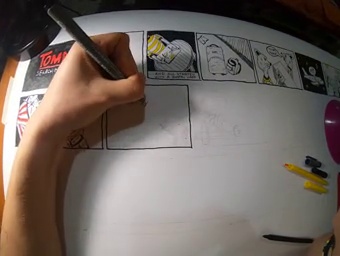}\vspace{4pt}
    \includegraphics[width=0.19\textwidth]{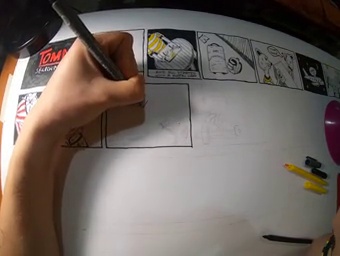}\vspace{4pt}
    \includegraphics[width=0.19\textwidth]{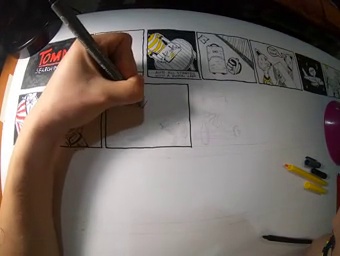}
    \end{minipage}
    }
    \subfigure[\texttt{\#C C stretches his left hand.}]{
    \begin{minipage}[]{0.9\textwidth}
    \includegraphics[width=0.19\textwidth]{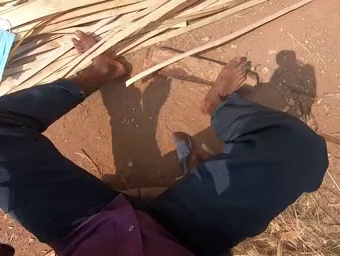}\vspace{4pt}
    \includegraphics[width=0.19\textwidth]{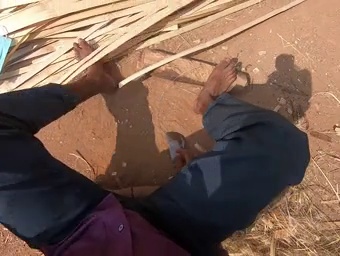}\vspace{4pt}
    \includegraphics[width=0.19\textwidth]{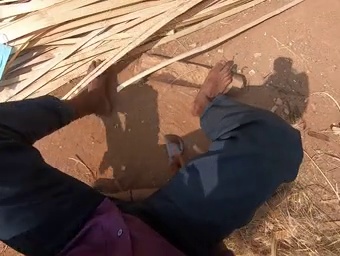}\vspace{4pt}
    \includegraphics[width=0.19\textwidth]{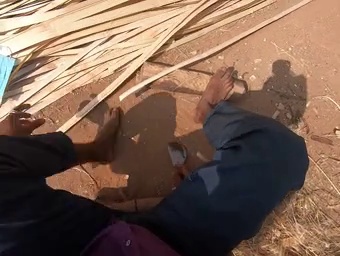}\vspace{4pt}
    \includegraphics[width=0.19\textwidth]{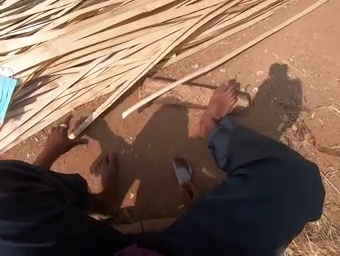}
    \end{minipage}
    }
    \subfigure[\texttt{\#O A man X moves hand from the table.}]{
    \begin{minipage}[]{0.9\textwidth}
    \includegraphics[width=0.19\textwidth]{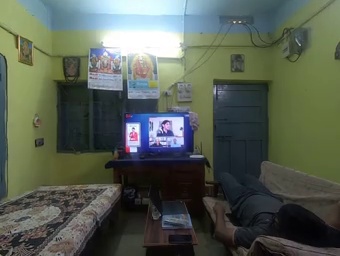}\vspace{4pt}
    \includegraphics[width=0.19\textwidth]{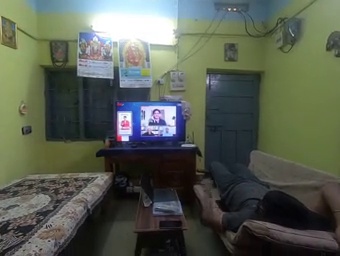}\vspace{4pt}
    \includegraphics[width=0.19\textwidth]{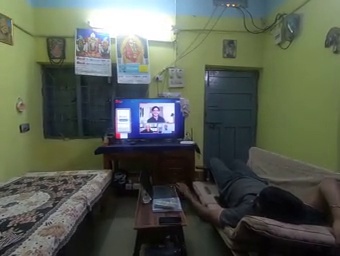}\vspace{4pt}
    \includegraphics[width=0.19\textwidth]{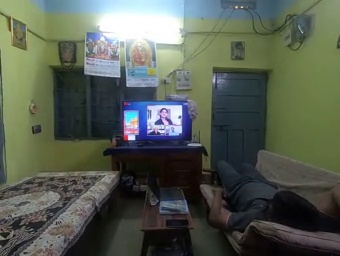}\vspace{4pt}
    \includegraphics[width=0.19\textwidth]{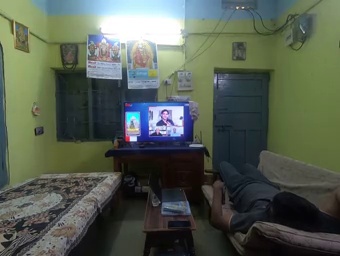}
    \end{minipage}
    }
    \caption{Visualization of \dataset~clip-text pairs. 
    We sample five frames uniformly for each clip and take its narration as its caption.}
    \label{fig_egoclip_vis}
    \vspace{-0.25cm}
\end{figure}

\noindent\textbf{Visualizations.}
In Fig.~\ref{fig_egoclip_vis}, we visualize some clip-text pairs created by our strategy. 
\begin{figure}[H]
\centering
\subfigure[Inter-video setting]{
\begin{minipage}[t]{1.0\textwidth}
\centering
\includegraphics[width=1.0\linewidth]{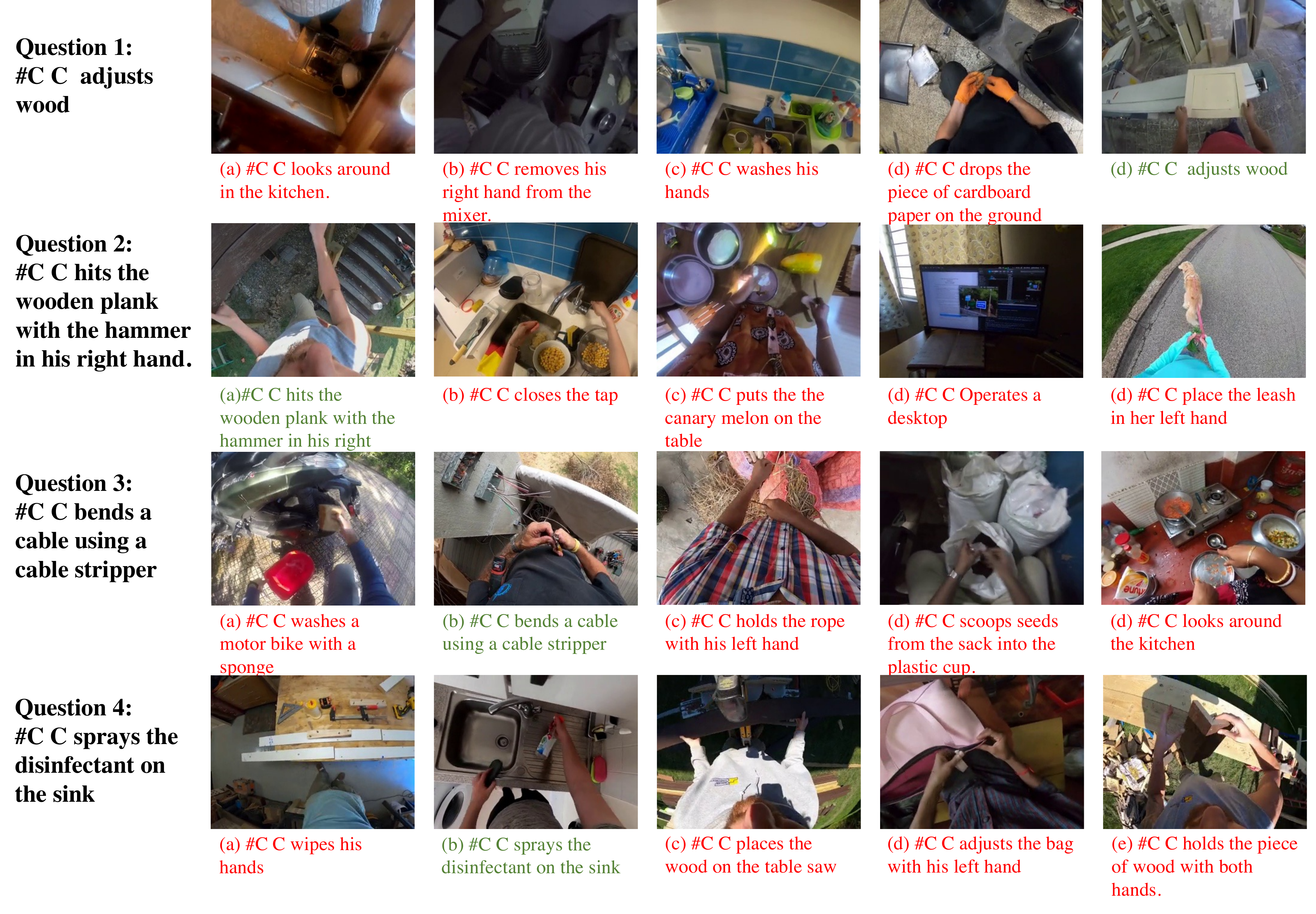}
\label{case_1}
\end{minipage}
}
\subfigure[Intra-video setting]{
\begin{minipage}[t]{1.0\textwidth}
\centering
\includegraphics[width=1.0\linewidth]{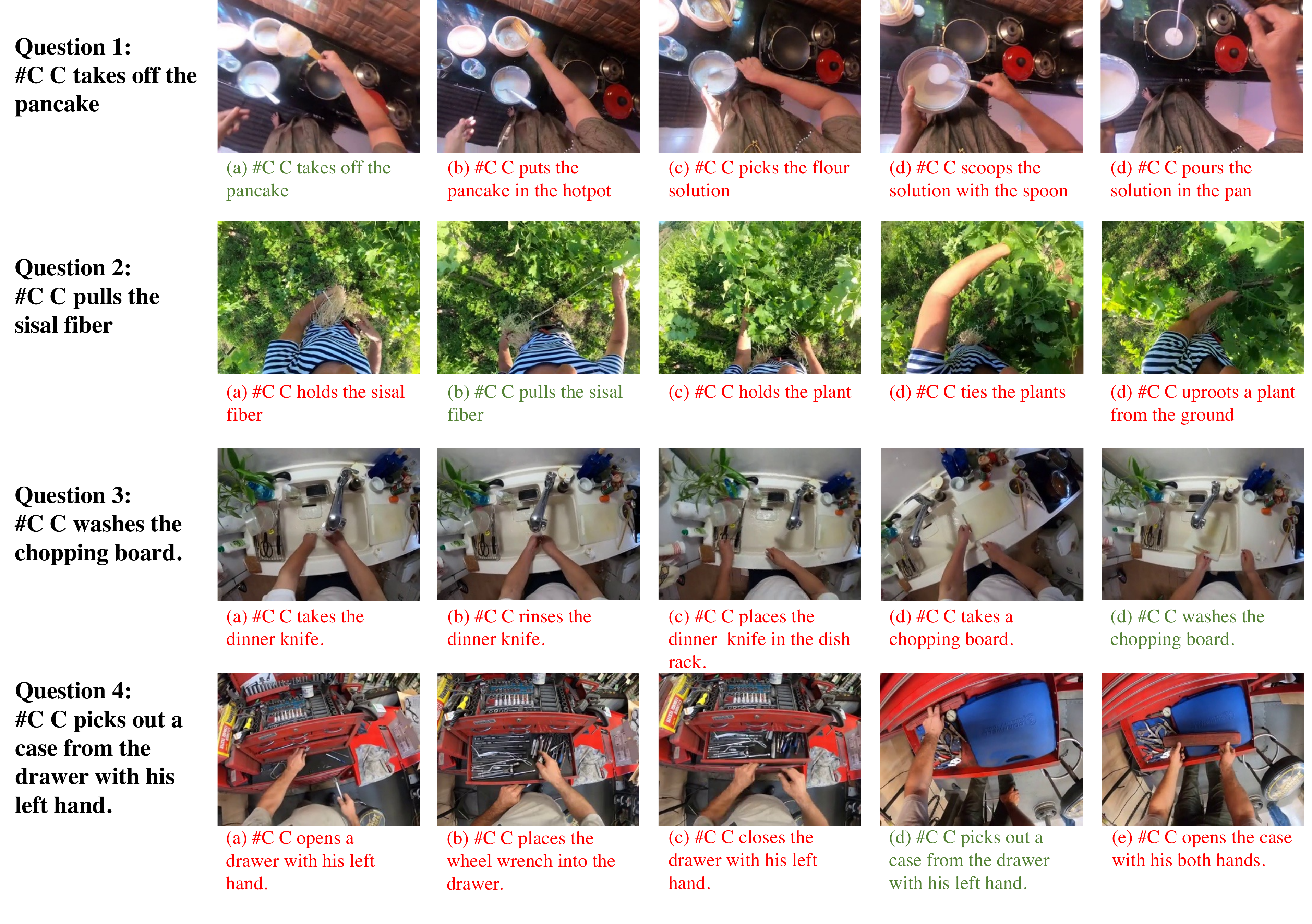}
\label{case_2}
\end{minipage}
}
\centering
\caption{Visualization of \eval~under two settings.
Left are the text questions; Right are the five candidate clips for each question and the text below as clip's narrations. 
The correct clip's narrations is highlighted in green and the wrong in red.}
\label{fig_egomcq_vis}
\end{figure}
\section{\secb}\label{b}
\subsection{Data deduplication}\label{b1}
To ensure repetitions do not appear in five options, we devise a deduplication strategy.
Initially, we use Sentence-BERT to extract sentence-level embeddings of narrations and set a manual threshold to remove repetitions.
But in this way, it is hard to control the fine-grained diversity between narrations, e.g., two narrations 
``\texttt{\#C C closes the refrigerator with his left hand.}'' and ``\texttt{\#C C opens the refrigerator with his left hand.}'' only differ in one word. These two sentences have a high score in sentence-level similarity, but are entirely different in semantic meanings. We hope to keep them and let the model distinguish them, especially in our intra-video setting.

Therefore, we propose to extract \textcolor{citecolor1}{the first verb} and \textcolor{citecolor2}{the first noun} of each narration and use them to define a tag for each narration.
The narrations shared with the same verb and the noun will be assigned the same tag. 
We also consider the words synonyms~(based on Ego4D taxonomy dictionary~\cite{grauman2021ego4d}).
For instance, ``\texttt{\#C C \textcolor{citecolor1}{take} the \textcolor{citecolor2}{phone}}'' and ``\texttt{\#C C \textcolor{citecolor1}{pick} the \textcolor{citecolor2}{cellphone}}'' are semantically same in verb and noun thus will be assigned the same tag. 
Then the narrations shared with the same tag are treated as repetitions, we only keep one of them and sample a new one until the tags of the five options are different.

\subsection{Multiple-views removing}\label{b3}
We first select videos from NUS/Minnesota/Georgia Tech/Indiana sources, which contribute to the multi-camera video data. Then, based on the metadata of the video (i.e. times when videos were collected), we observed that videos collected in the same timeframe tend to be multi-views of the same recording, so we manually group these videos into the same split to ensure the same scene does not appear in another split.

\subsection{Data analysis}\label{b2}
We finalize $39$K questions covering $198$K narrations with $468$ hours of video, where the ``inter-video'' has $24$K questions covering $290.3$ hours of videos. And the ``intra-video'' has $15$K questions and covers $178.3$ hours of videos. The average duration among the five options is $34.2$ seconds.

\begin{figure}[htb]
\vspace{-0.25cm}
    \centering
    \includegraphics[width=0.45\linewidth]{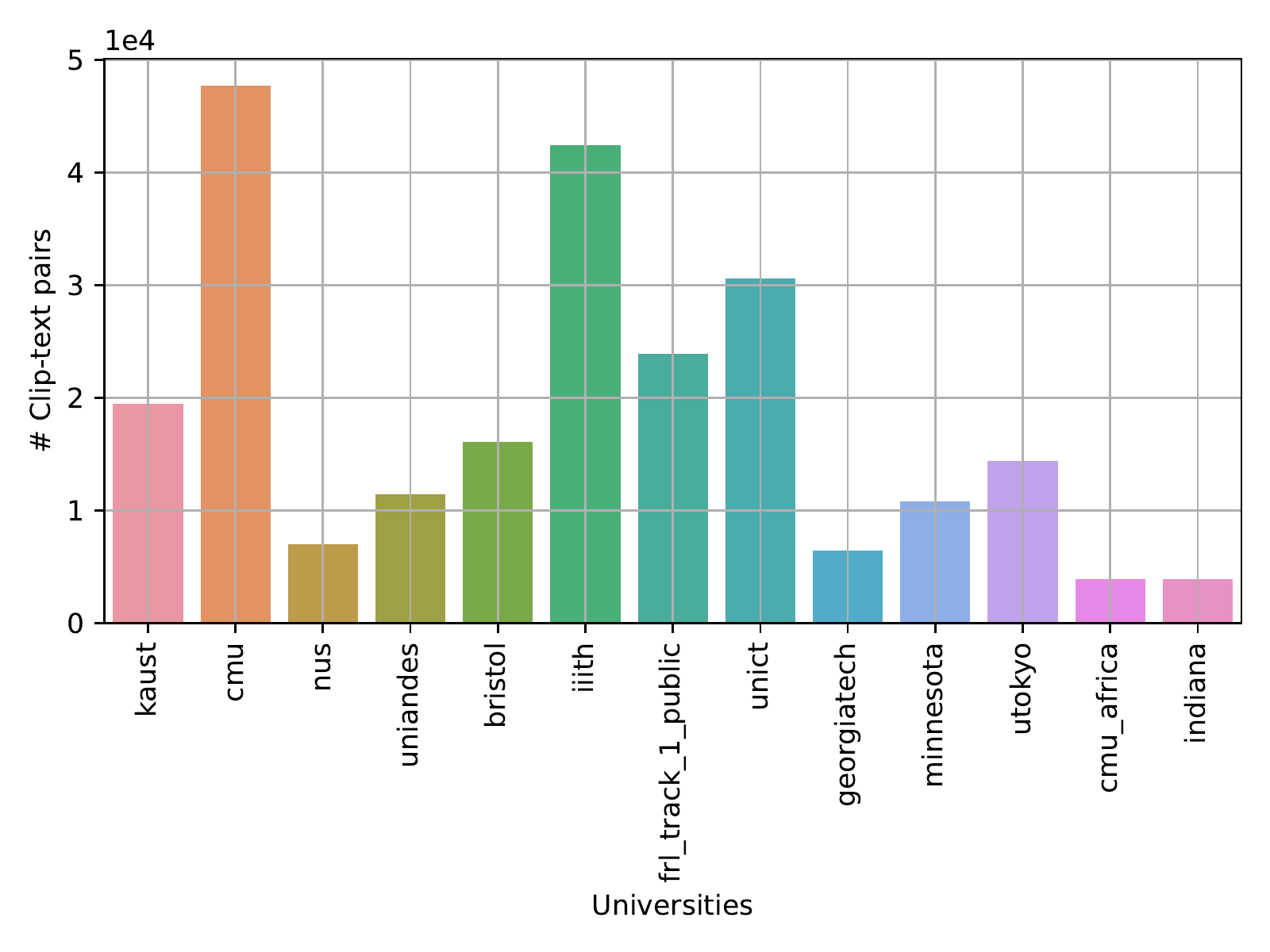}
    \caption{Institution distributions of \eval}
    \label{fig_egomcq_country}
\vspace{-0.25cm}
\end{figure}

\noindent\textbf{Geographic diversity.}
We present the geographic diversity of \eval~in Fig.~\ref{fig_egomcq_country}, which covers 13 institutions and is align with the geographic diversity of \dataset.

\noindent\textbf{Scenario diversity.}
In Fig.~5, we present the scenario distribution of \eval, which covers $74$ scenario. 
The largest scenario ``Cooking'' includes $49\text{K}~(15.3\%)$ clips and the smallest scenario ``Bus'' contains 6 instances.
\eval~covers $71\%$ of scenarios in \dataset and has other $3$ scenarios not appear in \dataset.
\eval~is close to \dataset~both in terms of geography and scene diversity, making it a good development set for \dataset~pretraining.

\noindent\textbf{Verbs and Nouns.}
\begin{figure}[htb]
\vspace{-0.25cm}
\centering
\subfigure[Top 50 most frequently verbs distribution]{
\begin{minipage}[t]{0.9\textwidth}
\centering
\includegraphics[width=1.0\linewidth]{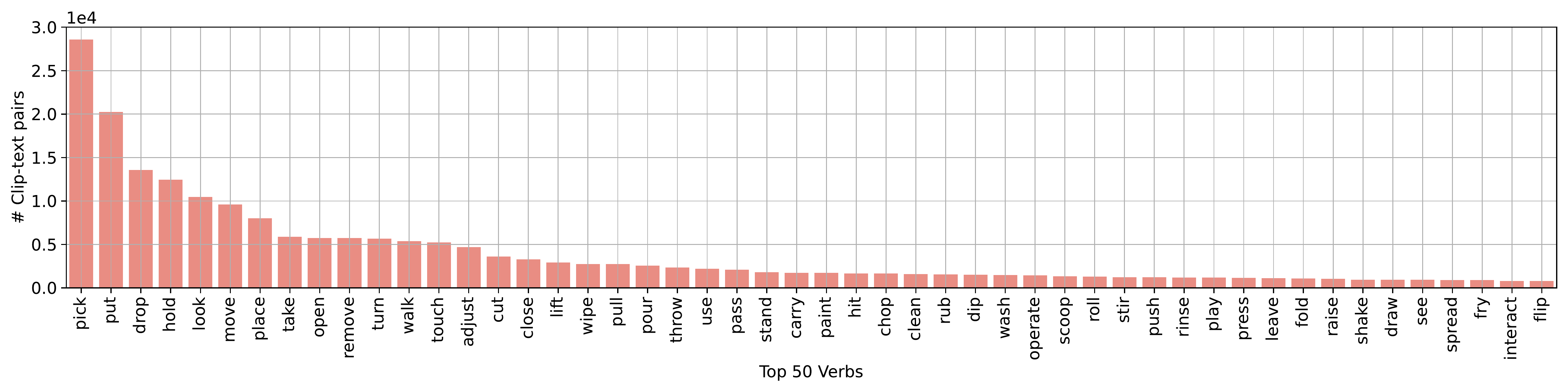}
\label{case_1}
\end{minipage}
}
\subfigure[Top 50 most frequently nouns distribution]{
\begin{minipage}[t]{0.9\textwidth}
\centering
\includegraphics[width=1.0\linewidth]{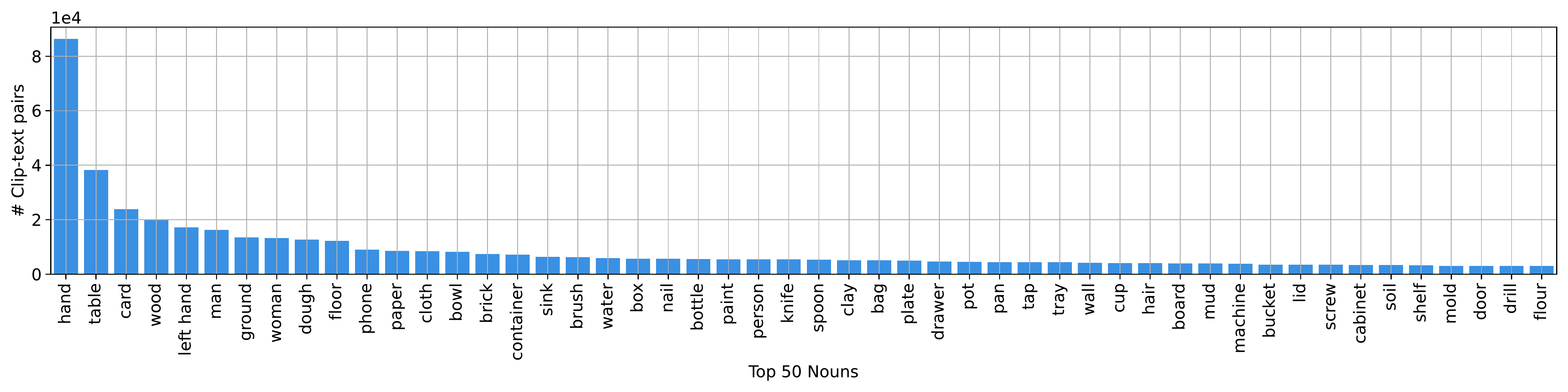}
\label{case_1}
\end{minipage}
}
\centering
\caption{Verbs and nouns distributions of \eval's narration.}
\label{fig_egomcq_v_n}
\vspace{-0.25cm}
\end{figure}
\eval~covers $198\text{K}$ narrations and each narration contains $3.15$ nouns and $0.97$ verbs in average. 
In Fig.~\ref{fig_egomcq_v_n}, we display the top 50 most frequently verb and nouns of \eval. The mostly common noun is ``hand'', covering $86\text{K}~(36.2\%)$ instances and the mostly frequently verb is ``pick'', which covers $28\text{K}~(12.0\%)$ instances.

\noindent\textbf{Visualization.}
In Fig.~\ref{fig_egomcq_vis}, we display examples of both the intra and inter settings of \eval.
\section{\secc}\label{c}
In this section, we present more technical details of our VLP model, mainly architecture and pretraining objective. 
\subsection{Architecture: Frozen-in-time~\cite{bain2021frozen}}
\textbf{Video Encoder.}
The video encoder is built upon Timesformer~\cite{timesformer}, a convolution-free video backbone that divides space-time attention in an efficient manner.
Take a RGB video clip $\mathcal{V}_i \in \mathbb{R}^{T\times3 \times H\times W}$ with $T$ frames and resolution $H\times W$ as input, the clip is first divided into $M\times N$ patches $\mathbf{p}\in \mathbb{R}^{M\times N\times 3\times P\times P}$ with size of $P\times P$, where $N=HW/P^2$. 
Next, patches $\mathbf{p}$ are linearly embed as a token sequence $\mathbf{z}\in \mathbb{R}^{MN\times D}$ with $D$ dimension. 
Then, the learned temporal embeddings $E^s\in \mathbb{R}^{N\times D}$ and spatial positional embeddings  $E^t\in \mathbb{R}^{N\times D}$ are added to each input token. Besides, a learnable $\texttt{CLS}$ token is concatenated at the beginning of the token sequence. 
Finally, these token sequences are fed into Timesformer and output the $\texttt{CLS}$ token of the last block, which is further projected to a $d$ dimension embedding by a linear layer to form the final clip representation $\mathbf{v}_i\in \mathbb{R}^{d}$.

\textbf{Text Encoder.}
The text encoder is built upon DistillBERT~\cite{distilbert}, which has $40\%$ less parameters than BERT while also preserves over $95\%$ performance, thus is efficient.
Taking a sentence $\mathcal{T}_i$ as input, first tokenize it as a sequence of tokens and feed it into DistillBERT. 
Similar to the video encoder, the $\texttt{CLS}$ token of DistillBERT's output is projected as $\mathbf{t}_i\in \mathbb{R}^d$ for the final text representation.

\subsection{Pretraining objective: \model}
To supplement the Eq.~\ref{nce} and Eq.~\ref{egonce}, we first formulate the complete form of InfoNCE:
\begin{equation}
\begin{aligned}
\mathcal{L}&=\mathcal{L}_{\text{v2t}}+\mathcal{L}_{\text{t2v}} \\
&=\frac{1}{|\mathcal{B}|}\sum_{i\in \mathcal{B}} \log \frac{\exp(\mathbf{v}_i^T\mathbf{t}_i /\tau)}{\sum_{j\in \mathcal{B}} \exp( \mathbf{v}_i^T\mathbf{t}_j /\tau)}+
\frac{1}{|\mathcal{B}|}\sum_{i\in \mathcal{B}} \log \frac{\exp(\mathbf{t}_i^T\mathbf{v}_i /\tau)}{\sum_{j\in \mathcal{B}} \exp( \mathbf{t}_i^T\mathbf{v}_j /\tau)}
\label{nce_dual}
\end{aligned}
\end{equation}

and our \model~extends the above as Eq.~\ref{egonce_dual} via two sampling strategies:

\begin{equation}
\begin{aligned}
	\mathcal{L}^\text{ego}&=\mathcal{L}_{\text{v2t}}^\text{ego}+\mathcal{L}_{\text{t2v}}^\text{ego}\\
	&=\frac{1}{| \mathcal{\widetilde{B}} |}\sum_{i\in\mathcal{\widetilde{B}}}  \log 
	\frac{
	{
	\sum_{k\in \mathcal{P}_i}\exp(\mathbf{v}_i^T\mathbf{t}_k /\tau)
	}
	}
	{  \sum_{j\in \mathcal{B}} \left( \exp(\mathbf{v}_i^T\mathbf{t}_j /\tau) +
	{\exp(\mathbf{v}_i^T\mathbf{t}_{j'} /\tau)}  \right) 
	} \\
	&+
	\frac{1}{| \mathcal{\widetilde{B}} |}\sum_{i\in\mathcal{\widetilde{B}}}  \log 
	\frac{
	{
	\sum_{k\in \mathcal{P}_i}\exp(\mathbf{t}_i^T\mathbf{v}_k /\tau)
	}
	}
	{  \sum_{j\in \mathcal{B}} \left( \exp(\mathbf{t}_i^T\mathbf{v}_j /\tau) +
	{\exp(\mathbf{t}_i^T\mathbf{v}_{j'} /\tau)}  \right) 
	}.
	\label{egonce_dual}
\end{aligned}
\end{equation}

For positive sampling~(the numerator term), we pre-extract the nouns and verbs for each narration $\mathcal{T}_i$ before pretraining and define two word vectors $\mathbf{w}_i^n\in \{0,1\}^{K_1}$ and  $\mathbf{w}_i^v\in \{0,1\}^{K_2}$ to encode the appearing nouns and verbs in sentence, where $K_1$ and $K_2$ denote the number of nouns and verbs in \dataset~(Refer to Sec~\ref{a}~narration analysis). 
During pretraining, for another instance $j$ within batch, we calculate the $s_{ij}=(\mathbf{w}_i^n)^T\mathbf{w}_j^n \cdot (\mathbf{w}_i^v)^T\mathbf{w}_j^v$, if $s_{ij}>0$, we regard instance $j$ is one of the positive sample $j\in \mathcal{P}_i$ of instance $i$.
Notably, the positive sampling space $\mathcal{P}$ would cover $\mathcal{\widetilde{B}}$ when working with the negative sampling strategy.

For negative sampling~(the denominator term), each time we sample an instance $i$, we sample an instance $i'\in \mathcal{V}_i$ in the same video and close in time~(less than 1 min) to generate the negative sample $i'\in \mathcal{N}(i)$ of instance $i$.
Notably, in this way, the actual instance within the batch $|\mathcal{\widetilde{B}}|=2N$ will be double the batch size $|\mathcal{B}|=N$.
In practice, we have to halve the batch size due to GPU memory limitations. 
Under halving the batch size, random sampling doesn't help in our method, which can be concluded by comparing baseline InfoNCE and variants (d) in Tab.~\ref{ablation_pos_neg} of the main body, where the batch size of the latter is half of the former. Despite this, our proposed sampling strategy~(f) can successfully improve the pretraining effect beyond baseline.

In contrast to the conventional negative sampling from the same video~\cite{wray2021semantic, lei2021detecting}, 
we specifically design our temporally adjacent negative sampling strategy to focus on the frequent appearance changes in egocentric videos, which has not been explored in previous approaches.
\section{\secd}\label{d}
\subsection{Implementation details}
Following the settings of official Frozen~\cite{bain2021frozen}, the video encoder is initialized with ViT~\cite{dosovitskiy2020image} weights trained on ImageNet-21K with sequence dimension $D=768$. 
The text encoder is based on huggingface's $\texttt{distilbert-base-uncased}$.
The dimension of common feature space is set as $256$, and the temperature parameter is set to $0.05$.
During pretraining, each video is resized to $224\times 224$ as input with sample frames number $4$ and batch size $512$.
We use the Adam optimizer with a learning rate of $3\times 10^{-5}$ with a total epoch of $10$. When transferring to downstream tasks, we select the checkpoints with the best score on \eval~benchmark i.e. average accuracy of inter-video and intra-video settings by default.

\subsection{Downstream settings}
We present the setting details of the downstream tasks we evaluated. 
For a fair comparison, for VLPs variants pretrained on different datasets, we use the same settings on downstream tasks, such as the fine-tuning objective.

\noindent\textbf{\epic~Multi-Instance Retrieval.}
In this task, after we finalize video-text pretraining, we continue to fine-tune the VLP model and keep most settings of pretraining~(e.g., input resolution, learning rate). 
Notably, we set the training epoch as $100$ and replace the training objective as Multi-instance Maxmargin loss in Eq.~\ref{mimm}, which is same as the baseline method JPoSE~\cite{wray2019fine}.
The reason for this is that in this task a narration may be jointly associated with multiple clips, so multi-instance learning mechanism can better handle such a situation. And this dataset also provides the action label to calculate the correlation $c_{ij}$ between two clip-text pairs $(i,j)$, which supports the implementation of Multi-instance Maxmargin loss.
\begin{align}
\mathcal{L}=
\sum_{(i,j,k)\in \Omega}\max
\left( \gamma + \mathbf{v}_i^T\mathbf{t}_j
-\mathbf{v}_i^T \mathbf{t}_k \right)
+
\left( \gamma + \mathbf{t}_i^T\mathbf{v}_j
-\mathbf{t}_i^T \mathbf{v}_k \right),
\label{mimm}
\end{align}
where $\Omega=\{(i,j,k)~|j\in i^{+}, k\in i^{-}\}$ is a triple, which indicates a positive instance $j$ and a negative instance $k$ for $i$. 
In our setting, we define the positive set as $i^{+}=\{j|c_{ij}>0.1\}$ and the negative as the remains sample within batch. The $\gamma$ is a margin factor and we set it as $0.2$.

\noindent\textbf{Charades-Ego Action Recognition.}
In this task, the textual categories are short phrases like ``Holding some clothes''. 
Thus, we regard this task as a kind of video-text retrieval by leveraging the text representation and using the InfoNCE as fine-tuning objective.
We set the epoch number as $10$ and keep other parameters unchanged.

\noindent\textbf{Ego4D \nlq}
This task is a kind of video-text localization and is hard to perform end-to-end training~(since a clip might long to $1200$ seconds).
The baseline method~\cite{zhang2020span} takes $2304$ dim SlowFast features~($1.87$ fps, with Kinetics 400 pretrained) and $768$ dim BERT features as input.
Therefore, we propose to replace the baseline input features as features of pretrained VLP video and text encoders to evaluate the pretraining effectiveness.
We extract the features with the same fps $1.87$ and sampling frame number $4$.
In fine-tuning stage, we keep the default setting of \cite{zhang2020span}.

\noindent\textbf{Ego4D \mq}
This task is a video-only task: temporal action localization. Similar to \nlq~task, we replace the input Slowfast features of baseline VSGN~\cite{zhao2021video} with VLP video features for evaluation. The extraction details are the same as \nlq.

\noindent\textbf{Ego4D \ossc}
This is an action classification task, we sample each clip with $16$ frames as input and use the cross-entropy as fine-tuning objective. The epoch is set as $10$.

\subsection{VLP Evaluation on \eval}
In Tab.~\ref{egomcq_eval}, we display \eval~evaluation result of Frozen pretrained on different video-text datasets.
\begin{table}[htb]
\centering
\vspace{-1em}
\begin{tabular}{l|cc}
    \toprule[1pt] 
	\multicolumn{1}{l|}{\multirow{2}{*}{\textbf{VL Pretraining}}} &  \multicolumn{2}{c}{\textbf{Accuracy~(\%)}} \\
	 & Intra-video      & Inter-video       \\ \midrule[1pt] 
	Random &  $20.0$ & $20.0$   \\ 	\midrule
	\epic &  $28.1$ & $22.7$   \\
    \howto &  $31.5$ & $21.6$   \\
	\ccweb  & $62.5$ & $27.4$   \\ \midrule
	\dataset &  $89.4$ & $51.5$   \\
    \dataset~w/~\model  &  $90.6$ & $57.2$   \\
	\bottomrule[1pt]
\end{tabular}
\vspace{0.4em}
\caption{Results of VLPs pretrained on different datasets in \eval}
\vspace{-2em}
\label{egomcq_eval}
\end{table}

As shown, pretraining with \epic~dataset~(1st-person view, $67.2\text{K}$ pairs) reach comparable performance with \howto~pretraining~(3rd-person view, $136\text{M}$ noisy pairs), which demonstrates the major domain gaps. 
Besides, Frozen with \ccweb~pretraining reach significant improvement on the intra-video setting, but minor in inter-video. 
We speculate this due to \ccweb~dataset covering a wide range of appearance information but still less exploration in the fine-grained action e.g. human-object interaction.

\subsection{Training Curves of \epic~video-text retrieval}
\begin{figure}[htb]
\centering
\vspace{-0.75cm}
\subfigure[mAP Curves]{
\begin{minipage}[t]{0.4\textwidth}
\centering
\includegraphics[width=1.0\linewidth]{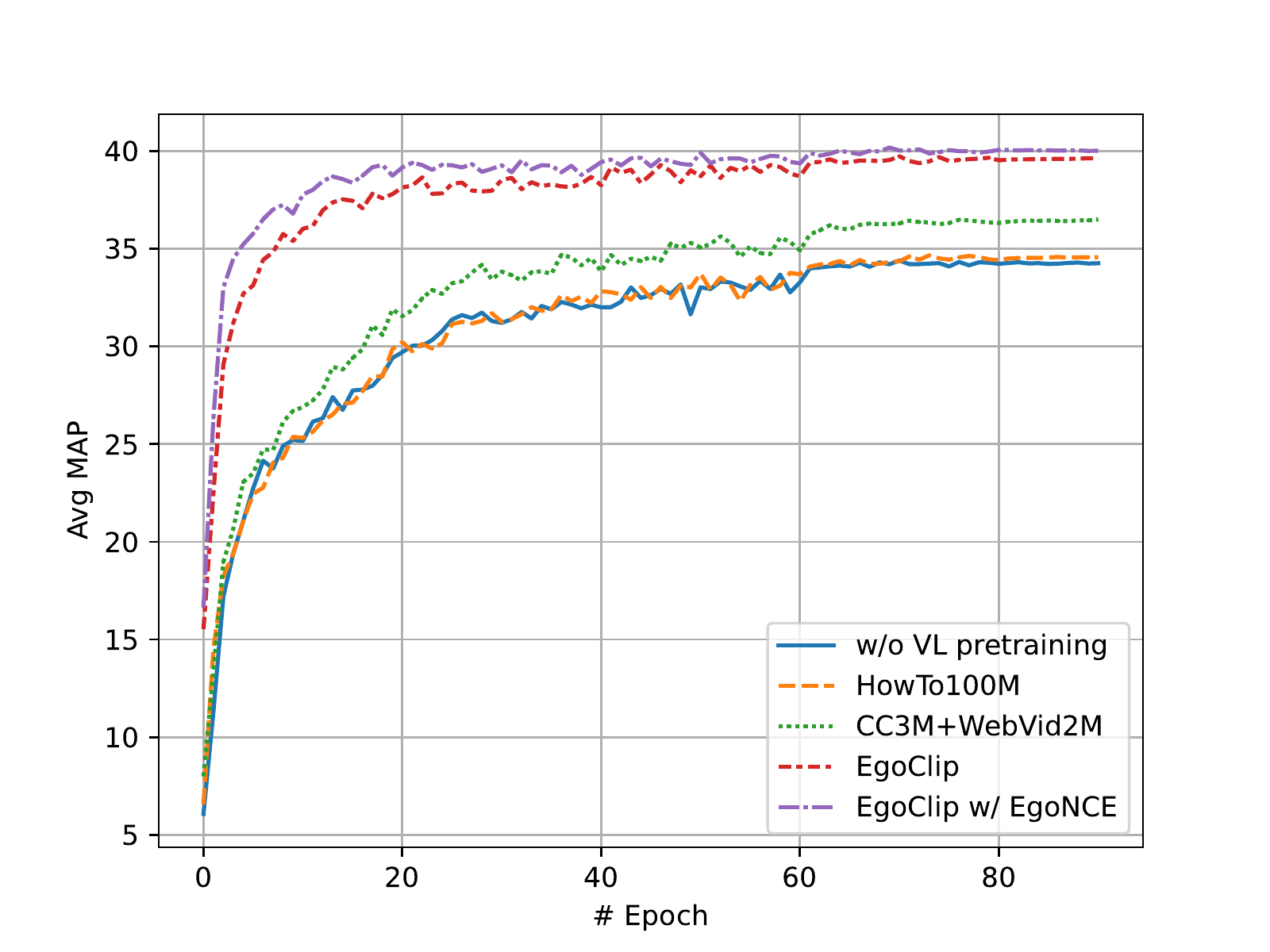}
\label{case_1}
\end{minipage}
}
\subfigure[nDCG Curves]{
\begin{minipage}[t]{0.4\textwidth}
\centering
\includegraphics[width=1.0\linewidth]{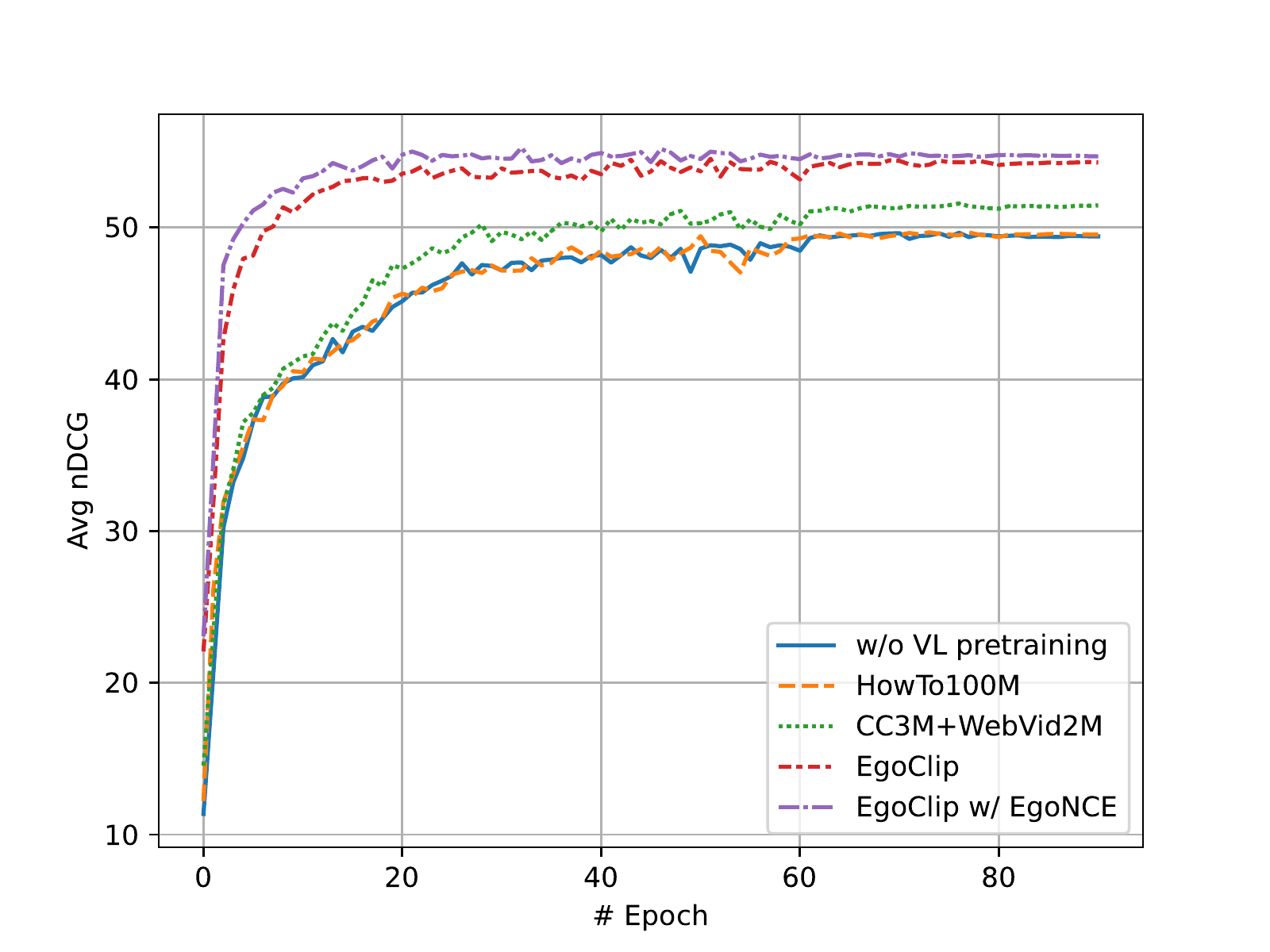}
\label{case_1}
\end{minipage}
}
\centering
\vspace{-0.25cm}
\caption{Training Curves of \epic~video-text retrieval}
\label{fig_epic_curve}
\end{figure}
In Fig.~\ref{fig_epic_curve}, we display training curves of \epic~video-text retrieval under different video-text pretraining, which also includes a baseline without video-text pre-training.
We can found that:
Variants with video-text pretraining have a faster rise in performance. 
Except for \howto, which is similar to variant without video-text pretraining. Especially with \dataset~for egocentric pretraining, the VLP model achieves nearly convergent performance with only a small number of epochs~(less than $20$).
With \model~as pretraining objective, this positive effect is further enhanced.

\subsection{Results on test set of \nlq}\label{d_nlq}
In Tab.~\ref{nlq_test}, we found the similar conclusions in test set of \nlq, pretraining with \dataset~and \model~reach the optimum performance.
\begin{table}[htb]
\centering
\begin{tabular}{clc|cccc}
	\toprule
	\textbf{Methods} & \multicolumn{2}{c|}{\textbf{Video-text Pre-extrated Features}} & \multicolumn{2}{c}{\textbf{IoU=0.3}} & \multicolumn{2}{c}{\textbf{IoU=0.5}} \\
	& Vis-text Enc & Vis-text PT & R@1 & R@5 & R@1 & R@5 \\  \midrule[1pt] 
	VSLNet  &  SlowFast+BERT   & - & $5.47$ & $11.21$ & $2.80$ & $6.57$  \\  
	 \midrule
 	VSLNet  &  Frozen       & \howto        & $3.77$ & $6.87$  & $1.62$ & $3.45$ \\	
	VSLNet  &  Frozen       & \ccweb & $4.87$ & $8.67$ & $2.50$ & $4.97$ \\		
	VSLNet  &  Frozen       & \dataset      & \underline{$10.34$} & \underline{$15.81$} & \underline{$6.24$} & \underline{$10.39$}\\
	VSLNet  &  Frozen+\model& \dataset      & $\mathbf{10.46}$ & $\mathbf{16.76}$ & $\mathbf{6.24}$ & $\mathbf{11.29}$ \\	
	\bottomrule
\end{tabular}
\centering
\vspace{0.4em}
\caption{Recall for several IoU on the NLQ task's test set.}
\label{nlq_test}
\vspace{-0.5cm}
\end{table}
\subsection{Results on test set of \mq}\label{d_mq}
We further display the test set results of \mq~in Tab.~\ref{mq_test}, pretraining with \dataset~and \model~reach the best performance, $3.78\%$ on R$@1$ and $4.65\%$ on Avg mAP over the baseline.
\begin{table}[htb]
\centering
{
\begin{tabular}{clc|cc}
	\toprule
	\textbf{Methods} & \multicolumn{2}{c|}{\textbf{Video-text Pre-extrated Features}} & \multicolumn{1}{c}{\textbf{IoU=0.5}} & \multicolumn{1}{c}{\textbf{mAP$(\%)$IoU}} \\
	& Vis-text Enc & Vis-text PT & R@1  & Avg \\  \midrule[1pt] 
	VSGN &  SlowFast & - & $24.25$  & $5.68$   \\ \midrule
	VSGN & Frozen & \howto&  $18.06$  & $5.28$   \\
	VSGN & Frozen & \ccweb  & $19.74$  & $5.95$   \\
	VSGN & Frozen & \dataset  & \underline{$27.98$}  & \underline{$9.78$}   \\
	VSGN & Frozen+\model & \dataset & $\mathbf{28.03}$  & $\mathbf{10.33}$   \\
	\bottomrule
\end{tabular}
}
\vspace{0.4em}
\centering
\caption{Recall and mAP metrics on the MQ task's test set.}
\label{mq_test}
\vspace{-0.5cm}
\end{table}

\subsection{Visualization}
\begin{figure}[htb]
\centering
\vspace{-0.3cm}
\centering
\includegraphics[width=1.0\linewidth]{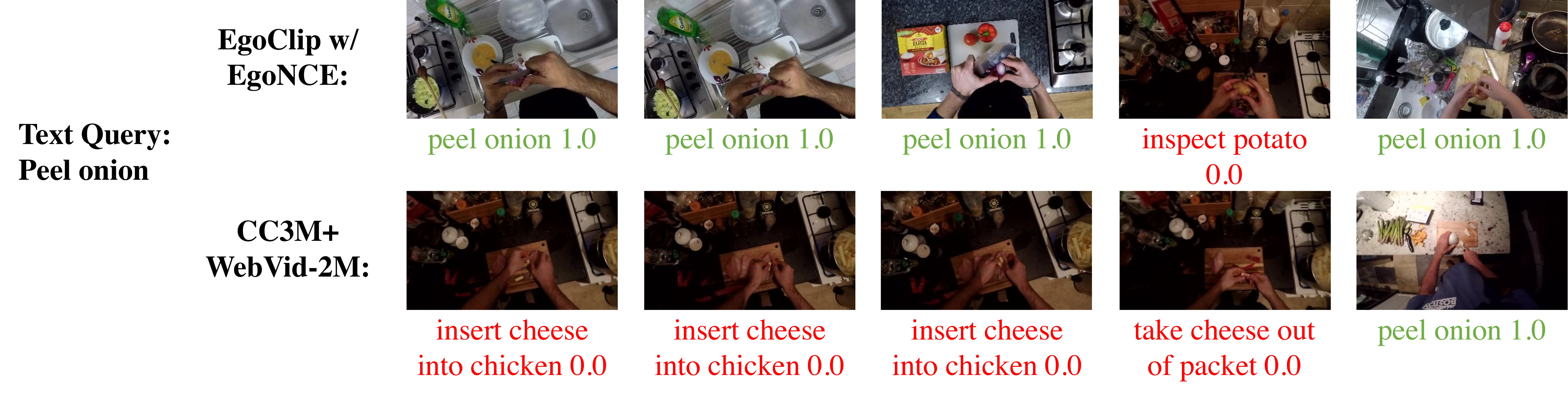}
\label{case_1}
\centering
\vspace{-0.4em}
\caption{Visualization of \epic~video-text retrieval. 
Given the same text query, we compare the \textbf{Top-5 results} of 1st-person pretraining and 3rd-person pretraining.}
\label{fig_epic_vis}
\end{figure}

To intuitively understand the effect of egocentric pre-training, in Fig.~\ref{fig_epic_vis}, we compare the \epic~video-text retrieval results between our pre-training~(\dataset~w/ \model) and \ccweb~pre-training, both fine-tuning with $16$ frames. 
The numbers after each narration represent the correlation scores between the query and the retrieval result, with $1$ being the best.

\end{document}